\documentclass{article}
\usepackage{iclr2027_conference,times}

\usepackage{amsmath,amssymb,amsfonts}
\usepackage{graphicx}
\usepackage{booktabs}
\usepackage{multirow}
\usepackage{array}
\usepackage{tabularx}
\usepackage[table]{xcolor}
\usepackage{caption}
\usepackage{placeins}
\usepackage{pifont}
\usepackage{hyperref}
\usepackage{url}

\hypersetup{
    pdftitle={Physical Time as a Missing Dimension of Robot Policy Memory Dynamics},
    pdfauthor={},
    pdfsubject={Robotics},
    pdfkeywords={Imitation Learning; Manipulation; Temporal Memory}
}

\providecommand{\vecsym}[1]{\mathbf{#1}}
\newcommand{\best}[1]{\textbf{#1}}
\colorlet{TableAccent}{black}
\colorlet{TableTint}{black!14}
\colorlet{TableMuted}{black!65}
\newcommand{\gain}[1]{\textcolor{TableAccent}{\ensuremath{+#1}}}
\newcommand{\gainlabel}[1]{\textcolor{TableAccent}{\textit{#1}}}
\newcommand{\ablscore}[2]{\shortstack[r]{#1\\[-1pt]{\scriptsize\textcolor{TableMuted}{\ensuremath{(#2)}}}}}
\newcommand{\tabnote}[1]{\par\vspace{3pt}{\footnotesize\raggedright #1\par}}

\newcommand{\paperauthorsplain}{Yushen Liang, Yue Peng, Baosheng Jin, Tianluo Zhang, Xinyu Zhang, Shuyi Zhou, Zhuoran Chen, Xinqi Liu, Shenji Wan}

\title{Physical Time as a Missing Dimension of Robot Policy Memory Dynamics}

\newcommand{\paperauthornames}{%
\parbox[t]{\dimexpr\textwidth-2\tabcolsep\relax}{%
\raggedright\bfseries
\mbox{Yushen Liang\textsuperscript{1,*}},
\mbox{Yue Peng\textsuperscript{1,*}},
\mbox{Baosheng Jin\textsuperscript{1,*}},
\mbox{Tianluo Zhang\textsuperscript{1}},
\mbox{Xinyu Zhang\textsuperscript{2}},
\mbox{Shuyi Zhou},
\mbox{Zhuoran Chen\textsuperscript{1}},
\mbox{Xinqi Liu\textsuperscript{1}},
\mbox{Shenji Wan\textsuperscript{1}}\par
}}
\author{%
\paperauthornames\\
\textsuperscript{1}New York University Shanghai, Shanghai, China\\
\textsuperscript{2}University of Electronic Science and Technology of China, Chengdu, China\\
\textsuperscript{*}Equal contribution
}

\usepackage{etoolbox}

\iclrfinalcopy
\makeatletter
\patchcmd{\@maketitle}
  {\lhead{Published as a conference paper at ICLR 2027}}
  {\lhead{}}
  {}
  {\PackageError{tfp-arxiv}{Could not remove the conference publication header}
    {Check the title macro in iclr2027_conference.sty before publishing.}}
\makeatother
\renewcommand{\headrulewidth}{0pt}
\hypersetup{hidelinks,pdfauthor={\paperauthorsplain}}
\begin{document}
\maketitle
\suppressfloats[t]

\begin{abstract}
Existing robot policy memories typically evolve with observation sequences, implicitly treating each update as the same amount of physical time. Yet identical observation histories can span different durations and require different actions, exposing physical time as a missing dimension of policy memory. We introduce Temporal Flow Policy (TFP), which learns memory dynamics to retain past events and evolve the memory state over measured time intervals. Observations and prior memory determine the content and time scales of each transition; elapsed time determines how far the state advances. TFP conditions diffusion and flow-matching policies through this memory state and is trained with their native imitation objectives. We evaluate TFP with Diffusion Policy and $\pi_{0.5}$ through controlled comparisons of memory designs. These experiments show that successful timing depends on both measured intervals and retained history. TFP achieves higher average success than the recurrent and state-space memories we equip with the same intervals and than a variant that feeds the interval to a learned gate. Memory readouts recover the time since an event and motion information after occlusion. Changing the supplied intervals systematically delays or advances actions, linking time inputs to behavior. Real-robot tasks requiring timing or memory, together with LIBERO evaluations, demonstrate benefits in physical and general manipulation. These results suggest viewing robot memory not as stored observation history, but as a state that retains past events and evolves with physical time.
\end{abstract}

\begin{figure}[t]
    \centering
    \includegraphics[width=\textwidth]{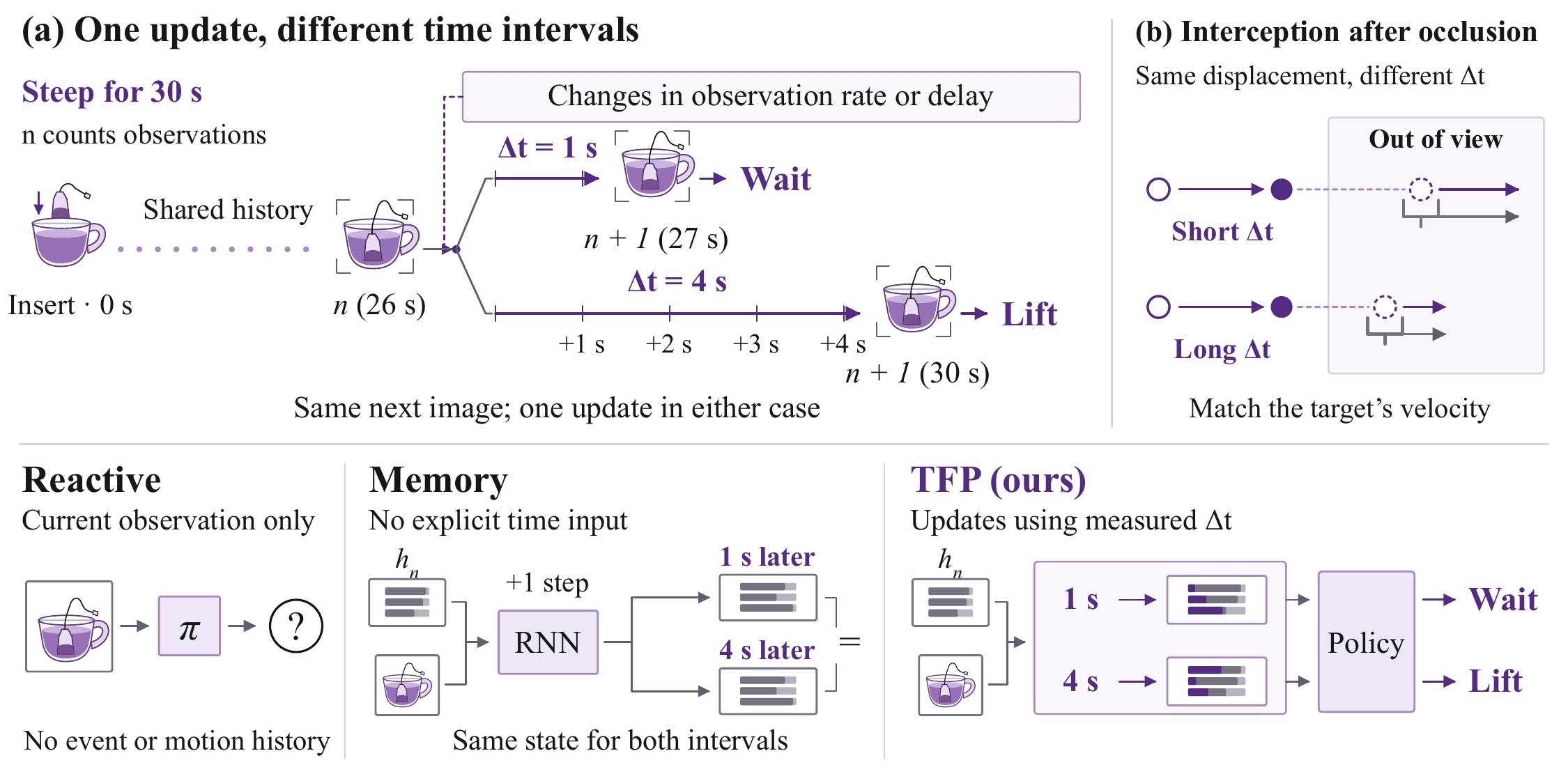}
    \caption{\textbf{Physical time gives remembered observations meaning for action.} (a) The same tea-steeping history and next image require waiting or lifting depending on elapsed time. (b) The same displacement over different intervals implies different velocities, which must be remembered for interception after occlusion. TFP uses measured intervals to evolve memory and guide action. Dashed paths indicate hidden motion; timings, trajectories, and memory states are schematic.}
    \label{fig:intro_overview}
\end{figure}

\section{Introduction}
\label{sec:introduction}

Consider a robot preparing tea. After it lowers an infuser into a cup, the scene may remain nearly unchanged, yet the correct action changes from waiting to lifting once the tea has steeped long enough. The robot needs to remember when steeping began and track how much time has passed. A robot picking up a cup from a moving conveyor faces a related challenge. It must estimate the cup's velocity from earlier observations and retain that information while the cup is briefly hidden from view. The same observed displacement over different time intervals implies different speeds, requiring the robot to match the cup's speed at contact. Policy memory must therefore use elapsed time both to respond to remembered events and to interpret motion from past observations (Figure~\ref{fig:intro_overview}).

Memory-augmented robot policies, including vision-language-action (VLA) models, retain earlier cues and track completed subtasks, allowing actions to depend on information beyond the current observation~\citep{koo2025hamlet,shi2025memoryvla,remem_vla2026}. Yet memories driven only by observation sequences treat each observation as one step: the same sequence can span different physical durations and require different actions. This mismatch becomes apparent when observation intervals vary with observation rate or inference latency. Timestamped histories make recording times available to the policy~\citep{yin2026simplememvla}. Continuous-time recurrent models provide a mechanism for evolving state over measured intervals~\citep{hasani2021ltc,hasani2022cfc}. We therefore study physical time as a missing dimension of robot policy memory: how remembered events and measured intervals jointly shape memory evolution and guide action.

We introduce \textbf{Temporal Flow Policy (TFP)}, which retains past events in a persistent memory state that evolves with physical time. At each query, separate heads use current observation features and previous memory to predict a candidate state and channel-wise time constants. A closed-form exponential update advances memory toward the candidate over the measured interval, so elapsed time directly governs state evolution. Prior memory shapes both predictions after visual cues disappear, while adaptive timescales support rapid updates around events and slower evolution during waiting. The updated memory conditions diffusion and flow-matching policies~\citep{chi2023diffusion,physicalintelligence2025pi05} through adaptive normalization, alongside their native observation and task context. We jointly learn memory dynamics and conditioning from causal demonstration histories using each backbone's native imitation objective, without event-time labels.

Controlled comparisons with Diffusion Policy and $\pi_{0.5}$ show higher average success than recurrent and state-space memories supplied with the same intervals and a learned-gate alternative. Memory readouts recover event age and motion information, while scaling the supplied intervals delays or advances actions without changing the physical execution schedule. Real-robot timing and sequencing tasks and LIBERO demonstrate benefits in physical and general manipulation.

Our contributions are:
\begin{itemize}
    \item We identify the ambiguity between observation order and elapsed physical time as a representation problem for robot policy memory.
    \item We introduce TFP, which uses learned event-dependent memory dynamics over measured intervals to condition diffusion and flow-matching policies.
    \item We connect temporal memory to behavior through time-aware controls, memory readouts, and clock interventions, and evaluate its utility in real-robot and benchmark tasks.
\end{itemize}

\section{Related Work}
\label{sec:related_work}

\paragraph{Memory in robot policies.}
Robot policies use history to recover information absent from the current
observation, including earlier cues and completed subtasks. HAMLET
aggregates compact moment tokens into memory features~\citep{koo2025hamlet},
while ReMem-VLA carries context through frame- and chunk-level recurrent
queries~\citep{remem_vla2026}. MemoryVLA retrieves perceptual and cognitive
representations from a consolidated memory bank~\citep{shi2025memoryvla},
and MEM combines short-term visual history with long-term language
memory~\citep{mem2026}. MemoryVLA uses episode timesteps to locate memory
entries, and MEM evaluates duration-sensitive cooking behavior. These systems
encode temporal context or evaluate timing behavior, but do not specify how
elapsed physical time should shape memory evolution.

\paragraph{Temporal information for robot control.}
The role of time depends on how it enters the policy. SimpleMemVLA timestamps
visual history relative to
the current decision~\citep{yin2026simplememvla}. DAM-VLA maintains
modality-specific latent buffers refreshed at different sensor rates,
decoupling action generation from individual encoding
schedules~\citep{vanjani2026damvla}. Time as a Control Dimension conditions
policies on remaining time and a time ratio that regulates the progression
of an internal clock~\citep{jia2026timecontrol}. These approaches make time
available through history, sensing schedules, or control conditions, but
leave open how elapsed time should enter memory updates: as an input
feature or an interval of state evolution.

\paragraph{Continuous-time dynamics in policy memory.}
Continuous-time recurrent models provide tools for modeling state evolution
over elapsed time. LTC learns adaptive time constants~\citep{hasani2021ltc}, while CfC provides
explicit time-dependent updates without numerical ODE
solvers~\citep{hasani2022cfc}. S4D discretizes continuous-time linear
state-space dynamics~\citep{gu2022s4d}, while Mamba learns input-dependent
discretization steps for selection~\citep{gu2023mamba}. M-CfC augments CfC with a gated memory pathway to support long-horizon object-goal navigation~\citep{chen2025mcfc}. TFP combines measured physical intervals with event-dependent memory dynamics to
condition diffusion and flow-matching action generators.

\section{Method}
\label{sec:method}

\subsection{Memory for Control in Physical Time}
\label{sec:temporal_interface}

\begin{figure}[!htb]
    \centering
    \includegraphics[width=\textwidth]{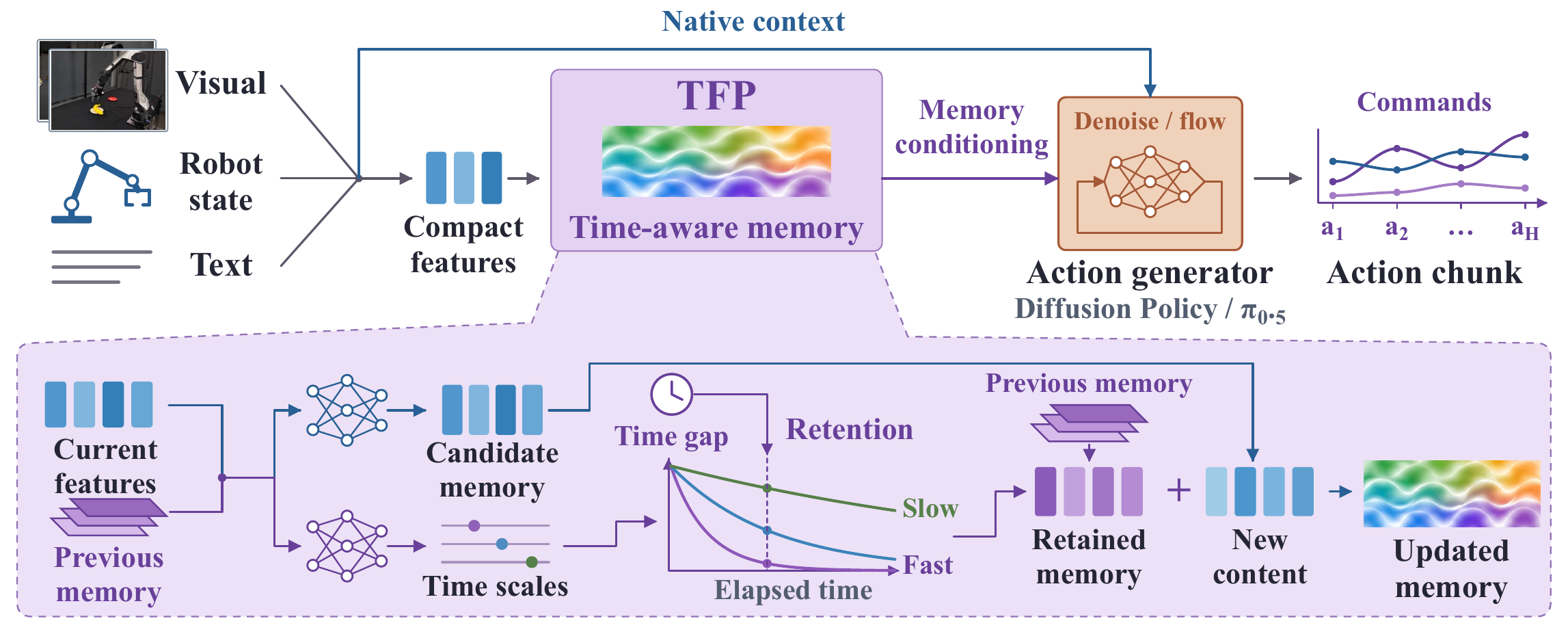}
    \caption{\textbf{Memory evolution and action generation in TFP.} Observation features and previous memory determine candidate content and adaptive time scales. The measured interval controls how far memory moves toward the candidate. The updated state conditions the backbone's action generator alongside its current context. Repeated previous-memory symbols denote the same state. Retention curves and command traces are schematic.}
    \label{fig:tfp_architecture}
\end{figure}

TFP combines past-event evidence and measured intervals in a persistent
memory state that conditions the action generator after events leave view
(Figure~\ref{fig:tfp_architecture}).

At query $n$, the backbone encodes observation $o_n$ and task specification
$g$ into compact features $\vecsym{x}_n$ and its native action context $C_n$.
The memory summarizes the timed history, while $C_n$ provides direct access
to the current scene and task:
\begin{equation}
\begin{aligned}
\vecsym{h}_n
&=\mathcal{U}_{\theta}(\vecsym{h}_{n-1},\vecsym{x}_n,\Delta_n),\\
\vecsym{A}_n
&\sim\pi_{\psi,b}(\,\cdot\mid C_n,\vecsym{h}_n).
\end{aligned}
\label{eq:tfp_interface}
\end{equation}
Here $\vecsym{h}_n\in\mathbb{R}^{d_h}$ is a persistent state,
$\vecsym{A}_n$ is an action chunk, and $b$ denotes the policy backbone.
For observations captured at timestamps $t_n$, $\Delta_n=t_n-t_{n-1}$ for $n\geq1$.
The memory state must infer how much time has passed since an event
from the preceding observations and intervals. Memory is
reset at episode boundaries; the first-query convention is specified in
Appendix~\ref{app:memory_parameterization}.

\subsection{Memory Dynamics in Physical Time}
\label{sec:liquid_memory}

\paragraph{Remembering past events.}
Following continuous-time recurrent models~\citep{hasani2021ltc,hasani2022cfc},
TFP predicts a candidate state
and positive, channel-wise time constants from the current features and
previous memory:
\begin{equation}
\begin{aligned}
\widehat{\vecsym{h}}_n
&=\tanh\!\left(f_{\theta}^{h}
  ([\vecsym{x}_n;\vecsym{h}_{n-1}])\right),\\
\vecsym{\tau}_n
&=f_{\theta}^{\tau}
  ([\vecsym{x}_n;\vecsym{h}_{n-1}])>\mathbf{0}.
\end{aligned}
\label{eq:memory_heads}
\end{equation}
TFP shares a local relaxation form with canonical LTC and parameterizes
candidate content and time scales through separate heads. Canonical LTC
couples these quantities through a shared drive
(Appendix~\ref{app:ltc_integration}). The separate heads let the model
learn what memory should represent and how quickly it changes, with past
events shaping both predictions through the prior state.

\paragraph{Evolving with elapsed time.}
Elapsed time sets how far memory moves toward the candidate:
\begin{equation}
\begin{aligned}
\vecsym{k}_n&=\exp(-\Delta_n/\vecsym{\tau}_n),\\
\vecsym{h}_n&=\vecsym{k}_n\odot\vecsym{h}_{n-1}
 +(1-\vecsym{k}_n)\odot\widehat{\vecsym{h}}_n.
\end{aligned}
\label{eq:memory_update}
\end{equation}
Operations are elementwise; $\vecsym{\tau}_n$ and $\Delta_n$ share physical units.

Unlike memories that receive elapsed time only as an additional feature,
TFP uses it as the integration interval of memory evolution. At fixed
observation and prior memory, $\Delta_n$ sets this interval without changing
the predicted candidate or time constants.

\paragraph{Local physical-time dynamics.}
The transition is the exact solution of a local relaxation process,
\begin{equation}
\frac{d\vecsym{y}(u)}{du}
=\frac{\widehat{\vecsym{h}}_n-\vecsym{y}(u)}{\vecsym{\tau}_n},
\qquad \vecsym{y}(0)=\vecsym{h}_{n-1}.
\label{eq:memory_local_dynamics}
\end{equation}
Fixing the predicted coefficients, the transition at
$u=\Delta_n$ preserves the state at zero duration, approaches the candidate
monotonically, and composes across intervals
(Appendix~\ref{app:time_consistency}). A generic learned gate need not
satisfy these constraints. The coefficients are recomputed at each query.

Under these fixed coefficients, a channel displaced from equilibrium
provides a monotonic temporal trace, with sensitivity decreasing toward
saturation. Different time constants provide complementary temporal
responses, allowing different memory channels to retain task-relevant events
over different physical timescales. Together, these channels form a
multi-timescale temporal basis for the memory state, echoing exponential
traces studied in neuroscientific models of temporal
representation~\citep{shankar2012time}. TFP's adaptive rates change with
events and interaction phases, supporting rapid updates at event onset
and slower evolution during waiting.

The exponential update specifies how a given memory transition depends on
elapsed time, including intervals absent from training. With the candidate
and time constants held fixed, changing the interval changes the extent of
relaxation through the same analytic rule. This provides a temporal
inductive bias for transferring from a single training cadence to different
observation schedules, which we evaluate through the controlled comparisons
in Section~\ref{sec:controlled_timing}.

\subsection{Acting on Memory}
\label{sec:memory_action}

Memory conditions action generation through
adaptive normalization~\citep{Peebles_2023_ICCV,perez2018film}. A learned
projection $M_b$ augments the native decoder condition:
\begin{equation}
\vecsym{c}_{n,s}
=\vecsym{c}^{\mathrm{native}}_{n,s}+M_b(\vecsym{h}_n).
\label{eq:memory_adapter}
\end{equation}
Here $s$ is diffusion or flow solver time; the condition modulates
normalized decoder activations.

Memory summarizes visual and robot-state features for both backbones.
With Diffusion Policy~\citep{chi2023diffusion}, it augments the denoiser's
adaptive layer-normalization condition. For $\pi_{0.5}$~\citep{physicalintelligence2025pi05},
it additionally summarizes language in controlled timing tasks, while LIBERO
retains language in the backbone context; both variants augment the action
expert's adaptive RMS-normalization condition.
Memory updates once per query and stays fixed during solver steps. Persistent state requires $\mathcal{O}(d_h)$ storage per episode, independent of episode length. The prediction heads, state update, and conditioning projection together cost $\mathcal{O}(d_h(d_x+d_h+d_c))$ per query, where $d_x$ and $d_c$ are feature and conditioning widths (Appendix~\ref{app:computational_overhead}).

\subsection{Learning Memory Dynamics from Demonstrations}
\label{sec:training}

The backbone's native imitation objective jointly trains memory dynamics
and the conditioning projection from demonstrated actions. Each action
loss uses the recurrent state computed from the causal
observation history and its measured intervals.
Controlled timing models reconstruct memory from causal prefixes;
LIBERO uses consecutive episode segments with gradients truncated
between segments. Appendices~\ref{app:eatb_details} and~\ref{app:timing_protocol}
detail sequence training and task protocols.

\section{Experiments}
\label{sec:experiments}

\begin{figure}[!htb]
    \centering
    \includegraphics[width=\textwidth]{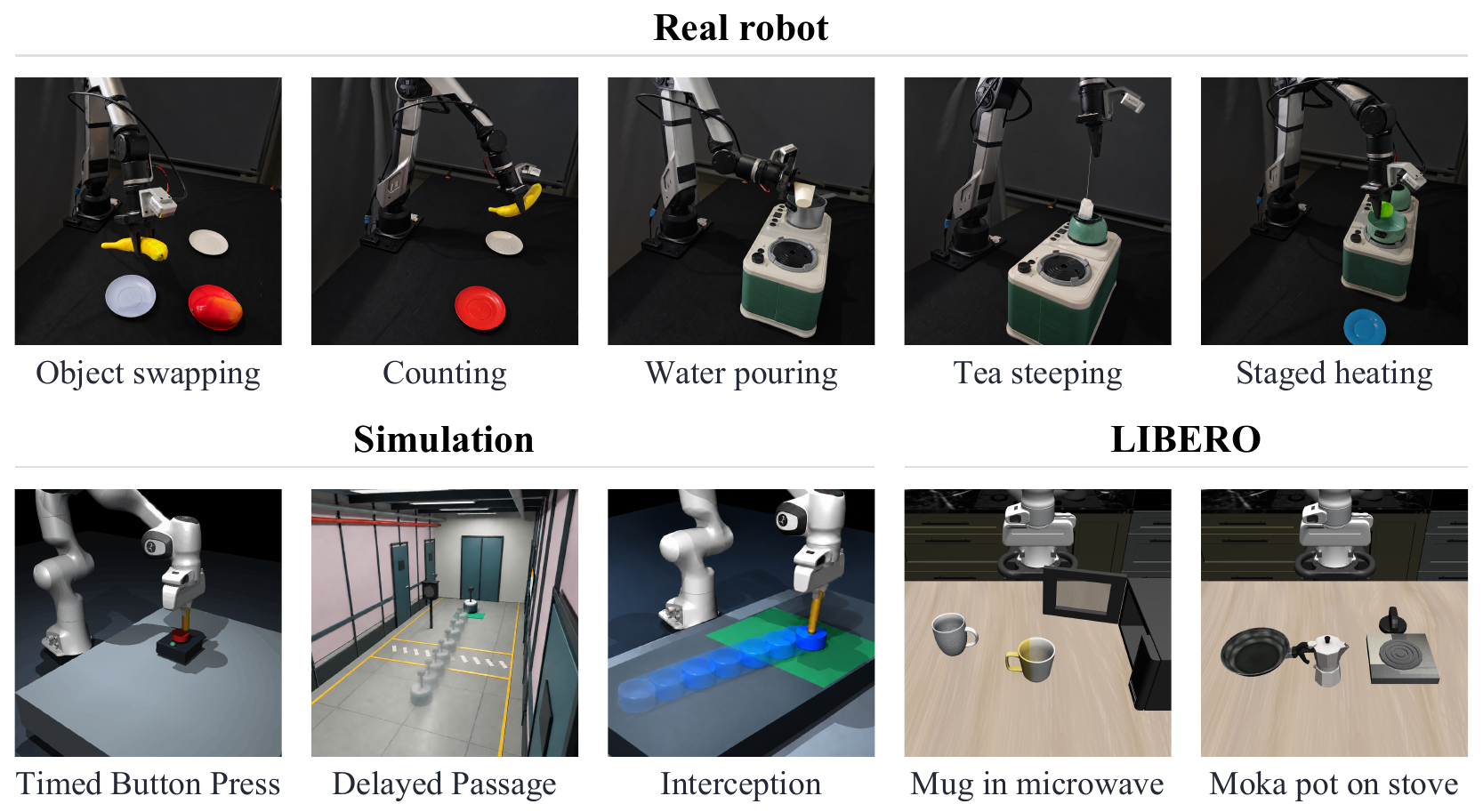}
    \caption{\textbf{Task overview.} Five real-robot tasks, three controlled simulation tasks, and two LIBERO~\citep{liu2023libero} tasks selected for illustration. LIBERO evaluation covers all 40 tasks across the four suites in Table~\ref{tab:libero_main_results}. Appendix~\ref{app:task_sequences} shows demonstration frames.}
    \label{fig:task_overview}
\end{figure}

We test control across observation schedules, connect temporal memory to action timing, and evaluate real-robot and general manipulation (Figure~\ref{fig:task_overview}).

\subsection{Does Policy Memory Need Physical Time?}
\label{sec:controlled_timing}

\paragraph{Separating duration from observation count.}
Both Diffusion Policy and $\pi_{0.5}$ models are trained with fixed 0.1\,s observation intervals, without mixed-cadence sampling augmentation. Time-aware models receive the measured observation interval: it is 0.1\,s during training and follows the actual observation schedule at evaluation. Each trained model is evaluated at 10\,Hz, 5\,Hz, and under jitter and dropout, with actions continuing between queries. Task durations and target speeds remain fixed, so equal physical durations can span different observation counts.

\textbf{Timed Button Press} requires continuous contact for 3.0--3.5 seconds before release and retraction. \textbf{Delayed Passage} presents a brief cue at a randomized onset and requires crossing 6--9 seconds later. \textbf{Interception} requires matching a previously observed target's velocity after occlusion; the robot moves only after the target disappears. These tasks test action duration, event age, and motion estimation. Time since contact, cue timestamps, and target velocity are not policy inputs.

\paragraph{Evaluation.}
We evaluate Diffusion Policy~\citep{chi2023diffusion} and $\pi_{0.5}$~\citep{physicalintelligence2025pi05} with reactive and short-history controls, and GRU~\citep{cho2014learning} and S4D~\citep{gu2022s4d} both with and without measured intervals. We also evaluate LTC~\citep{hasani2021ltc} with measured intervals for both backbones. The $+\Delta t$ suffix denotes measured intervals. We report mean success across three training seeds. Per seed and schedule, we evaluate 50 trials each for Button Press and Passage and 300 direction--speed cases for Interception, paired across methods within each backbone. Appendices~\ref{app:timing_protocol} and~\ref{app:timing_uncertainty} give protocols and task-average sample SDs.

\begin{table}[!htb]
\centering
\caption{\textbf{Temporal control across observation schedules.} Mean success over three training seeds (\%, $\uparrow$). Fixed-$\Delta t$ uses $0.1$\,s at inference. Bold: column maxima by backbone, including ties.}
\label{tab:timing_main_results}
\small
\setlength{\tabcolsep}{2pt}
\renewcommand{\arraystretch}{1.25}
\begin{tabularx}{\textwidth}{@{}l*{12}{>{\raggedleft\arraybackslash}X}@{}}
\toprule
 & \multicolumn{4}{c}{\textbf{Button Press}} & \multicolumn{4}{c}{\textbf{Passage}} & \multicolumn{4}{c}{\textbf{Interception}} \\
\cmidrule(lr){2-5}\cmidrule(lr){6-9}\cmidrule(lr){10-13}
Memory & 10 Hz & 5 Hz & Jit. & Drop. & 10 Hz & 5 Hz & Jit. & Drop. & 10 Hz & 5 Hz & Jit. & Drop. \\
\midrule
\multicolumn{13}{l}{\emph{Diffusion Policy}} \\
Reactive & 0.0 & 0.0 & 0.0 & 0.0 & 0.0 & 0.7 & 5.3 & 3.3 & 8.7 & 10.3 & 11.7 & 11.7 \\
History & 0.0 & 0.0 & 0.0 & 0.0 & 1.3 & 2.7 & 2.7 & 2.0 & 30.0 & 34.3 & 29.0 & 23.0 \\
GRU & 42.7 & 4.7 & 28.0 & 3.3 & 92.0 & 4.0 & 50.0 & 5.3 & 40.3 & 10.3 & 32.3 & 18.3 \\
GRU+$\Delta t$ & 44.0 & 5.3 & 30.0 & 4.7 & 90.0 & 5.3 & 50.0 & 6.0 & 41.3 & 11.0 & 33.0 & 19.0 \\
S4D & 52.7 & 16.0 & 24.0 & 10.0 & 90.0 & 12.0 & 44.0 & 16.0 & 36.3 & 16.3 & 26.3 & 20.3 \\
LTC+$\Delta t$ & 18.7 & 13.3 & 20.7 & 11.3 & 24.7 & 18.0 & 22.7 & 16.0 & 24.3 & 21.1 & 25.4 & 20.0 \\
S4D+$\Delta t$ & 56.7 & 46.0 & 50.0 & 40.0 & 96.7 & 94.7 & 96.7 & 92.7 & 40.3 & 46.3 & 44.3 & 40.3 \\
TFP fixed $\Delta t$ & \best{88.0} & 0.7 & 12.7 & 0.7 & \best{97.3} & 0.7 & 32.7 & 0.7 & \best{55.3} & 15.9 & 39.1 & 17.4 \\
\addlinespace[2pt]
\rowcolor{TableTint}
\textbf{TFP} \textit{(ours)} & \best{88.0} & \best{94.7} & \best{76.0} & \best{66.0} & \best{97.3} & \best{96.7} & \best{98.7} & \best{96.7} & \best{55.3} & \best{52.6} & \best{56.3} & \best{51.3} \\
\midrule
\multicolumn{13}{l}{\emph{$\pi_{0.5}$}} \\
Reactive & 0.0 & 0.0 & 0.0 & 0.0 & 6.7 & 16.0 & 8.7 & 6.0 & 14.0 & 18.4 & 18.1 & 22.6 \\
History & 0.0 & 14.0 & 4.7 & 3.3 & 2.7 & 4.7 & 0.7 & 3.3 & 35.6 & \best{35.1} & 35.1 & 34.7 \\
GRU & 6.7 & 1.3 & 28.0 & 6.0 & \best{90.0} & 4.7 & 46.0 & 6.7 & 31.3 & 18.3 & 30.3 & 28.3 \\
GRU+$\Delta t$ & 7.3 & 2.7 & 30.0 & 6.7 & 88.0 & 5.3 & 48.0 & 8.0 & 32.3 & 19.0 & 31.0 & 29.0 \\
S4D & 90.0 & 10.0 & 20.0 & 8.0 & 52.7 & 10.0 & 24.0 & 12.0 & 28.3 & 10.3 & 20.3 & 16.3 \\
LTC+$\Delta t$ & 23.3 & 18.7 & 22.7 & 16.0 & 26.7 & 22.7 & 28.7 & 18.7 & 29.3 & 25.3 & 28.3 & 23.3 \\
S4D+$\Delta t$ & 92.0 & \best{98.7} & 93.3 & 88.0 & 52.0 & 84.0 & 78.0 & \best{84.0} & 29.1 & 32.4 & 34.0 & 35.6 \\
TFP fixed $\Delta t$ & \best{94.0} & 0.7 & 14.0 & 0.7 & \best{90.0} & 0.7 & 46.0 & 6.0 & \best{47.3} & 20.6 & 34.3 & 26.4 \\
\addlinespace[2pt]
\rowcolor{TableTint}
\textbf{TFP} \textit{(ours)} & \best{94.0} & 95.3 & \best{94.7} & \best{90.0} & \best{90.0} & \best{84.7} & \best{88.0} & \best{84.0} & \best{47.3} & 33.2 & \best{37.1} & \best{41.0} \\
\bottomrule
\end{tabularx}
\tabnote{The $+\Delta t$ suffix denotes measured intervals; plain GRU and S4D are independently trained without them for both backbones. Jit./Drop.: jitter/dropout. Task-average sample SDs: Appendix~\ref{app:timing_uncertainty}.}
\end{table}

\paragraph{Memory improves control.}
Reactive policies fail Button Press under every schedule with both backbones (Table~\ref{tab:timing_main_results}). Short history improves Interception but achieves at most 4.7\% on Passage, where recurrent memory must retain the cue throughout the waiting interval.

Averaged equally across tasks and schedules, TFP achieves the highest reported success: 77.5\% with Diffusion Policy versus 62.1\% for the stronger time-aware control, S4D+$\Delta t$; with $\pi_{0.5}$, the corresponding rates are 73.3\% and 66.8\%. Passage performance remains high across schedules, supporting responses to remembered cues despite changes in observation cadence.

\subsection{How Memory Controls Behavior}
\label{sec:ablations}
\label{sec:clock_interventions}
\label{sec:memory_representations}

Frozen-weight interventions test how clock inputs and stored memory affect behavior; separately trained ablations assess architectural choices. Linear probes read out information from memory.

\begin{figure}[!htb]
    \centering
    \includegraphics[width=\textwidth]{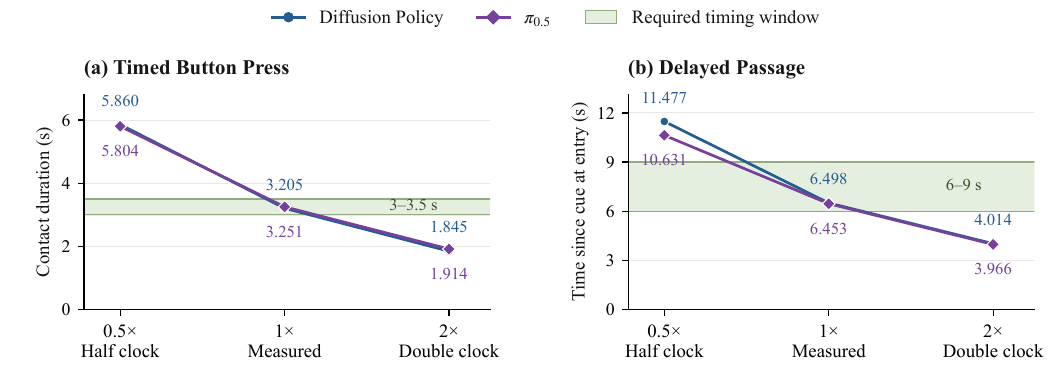}
    \caption{\textbf{Scaling the supplied clock shifts action timing.} Median physical durations under jitter versus supplied interval multiplier $\alpha$ (inputs $\alpha\Delta_n$); weights and physical query schedules stay fixed. Shading marks the required timing windows. Button measures contact until release; Passage measures first attempted entry since cue onset. Medians use observed events: DP Button $n=20/38/36$ of 50 paired trials for $\alpha=0.5/1/2$; all other points $n=50$.}
    \label{fig:clock_scaling}
\end{figure}

\paragraph{Action timing follows the supplied clock.}
We scale the intervals supplied to trained policies while keeping weights and physical execution schedules fixed. Under jitter, halving the intervals delays button release and passage entry, whereas doubling them advances these actions (Figure~\ref{fig:clock_scaling}). For $\pi_{0.5}$ on Button Press and both backbones on Passage, each directional shift occurs in all 50 paired trials. Both scalings eliminate success on these tasks with either backbone (Appendix~\ref{app:clock_interventions}).

Replacing measured intervals with 0.1\,s reduces success, and clearing memory at every query eliminates success on all three tasks at 10\,Hz (Appendix Table~\ref{tab:core_interventions}). Timing thus depends on both measured intervals and accumulated history.

\paragraph{Elapsed event time is recoverable from memory.}
Linear probes recover event duration and motion from frozen Diffusion Policy's TFP memory. For Passage, predicting time since the cue from memory reduces mean absolute error from about 0.83\,s using episode time alone to 0.04\,s, supporting event-relative timing beyond episode progress. After occlusion, memory also retains speed information beyond single-frame cues. Appendix~\ref{app:temporal_readout} details these comparisons and protocols.

\paragraph{Memory conditioning supports action generation.}
\label{sec:action_conditioning}
Without memory conditioning throughout Diffusion Policy training and evaluation, success is 0.5\% on Button Press and 3.7\% on Passage, compared with the Full TFP means of 81.2\% and 97.3\%, respectively (Appendix~\ref{app:memory_retention}).

\begin{samepage}
\paragraph{Three design choices for physical-time memory.}
Separately trained ablations remove history from the candidate, use static time constants, or replace interval-based evolution with a learned $\Delta t$ gate (Table~\ref{tab:core_structure_ablations}). Each lowers mean success relative to Full TFP, supporting event-dependent content, adaptive timescales, and the update rule in Section~\ref{sec:liquid_memory}. On Diffusion Policy Button and Passage, the learned gate achieves 14.5\% and 15.0\%, compared with 81.2\% and 97.3\% for TFP's exponential update. Appendix~\ref{app:tau_ablation} gives additional ablations.
\par\end{samepage}

\begin{table}[!htb]
\centering
\caption{\textbf{Which memory components matter?} Mean success over three training seeds (\%, $\uparrow$), averaged equally over four schedules. Parentheses show changes from Full TFP (pp).}
\label{tab:core_structure_ablations}
\small
\setlength{\tabcolsep}{3pt}
\renewcommand{\arraystretch}{1.18}
\begin{tabularx}{\textwidth}{@{}l*{6}{>{\raggedleft\arraybackslash}X}@{}}
\toprule
 & \multicolumn{3}{c}{Diffusion Policy} & \multicolumn{3}{c}{$\pi_{0.5}$} \\
\cmidrule(lr){2-4}\cmidrule(lr){5-7}
Variant & Button & Passage & Intercept. & Button & Passage & Intercept. \\
\midrule
\rowcolor{TableTint}
\textbf{Full TFP} \textit{(ours)} & \best{81.2} & \best{97.3} & \best{53.9} & \best{93.5} & \best{86.7} & \best{39.7} \\
Candidate w/o history & \ablscore{42.7}{-38.5} & \ablscore{8.0}{-89.3} & \ablscore{44.2}{-9.7} & \ablscore{12.0}{-81.5} & \ablscore{65.7}{-21.0} & \ablscore{31.7}{-8.0} \\
Static time constants & \ablscore{58.5}{-22.7} & \ablscore{82.7}{-14.6} & \ablscore{44.8}{-9.1} & \ablscore{64.3}{-29.2} & \ablscore{13.3}{-73.4} & \ablscore{31.4}{-8.3} \\
Learned $\Delta t$ gate & \ablscore{14.5}{-66.7} & \ablscore{15.0}{-82.3} & \ablscore{30.5}{-23.4} & \ablscore{27.0}{-66.5} & \ablscore{37.0}{-49.7} & \ablscore{30.1}{-9.6} \\
\bottomrule
\end{tabularx}
\tabnote{Task-average sample standard deviations across the three seeds appear in Appendix~\ref{app:timing_uncertainty}.}
\end{table}

\begin{figure}[!htb]
    \centering
    \includegraphics[width=\textwidth]{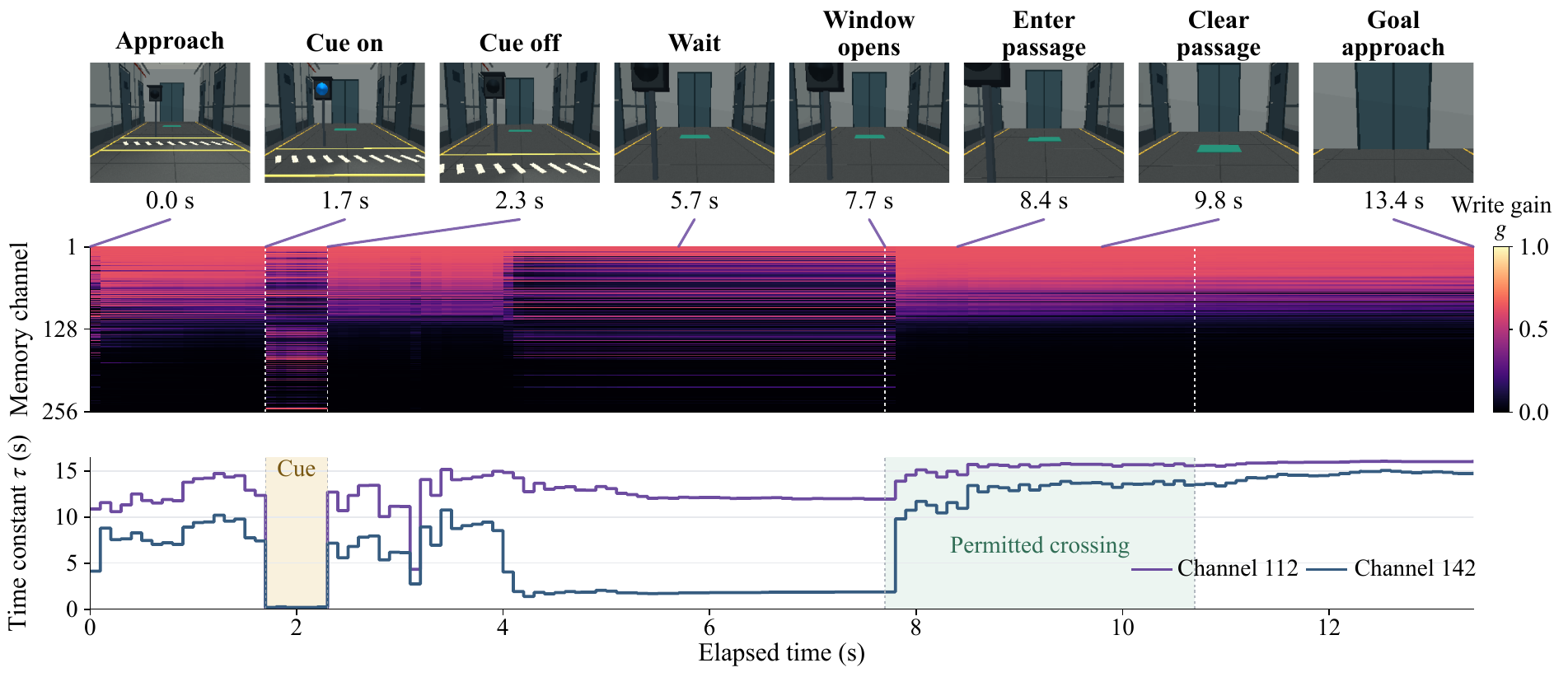}
    \caption{\textbf{Memory dynamics during Delayed Passage.} Observations, channel-wise write gains, and two channel time constants share a physical time axis. Heatmap cells and step traces start at their query timestamps. Shading marks the cue and permitted crossing window. A demonstration is replayed through frozen $\pi_{0.5}$ with TFP; Appendix~\ref{app:temporal_readout} details channel and trajectory selection.}
    \label{fig:navigation_memory_dynamics}
\end{figure}

Figure~\ref{fig:navigation_memory_dynamics} illustrates these dynamics around the passage cue. Different channels exhibit distinct temporal responses, with some rapidly adapting around events and others maintaining slower traces during waiting. Write gains, $\vecsym{g}_n=\mathbf{1}-\vecsym{k}_n$, show the resulting variation across channels. Appendix~\ref{app:simulation_memory_visualizations} shows corresponding event-aligned visualizations for Button Press and Interception.

\subsection{Does TFP Improve Real-Robot Control?}
\label{sec:real_world}

On a Galaxea A1, we compare TFP with each backbone's standard reactive policy. Object swapping, counting pick-and-place, and water pouring require tracking action sequences. Tea steeping requires $3.0\pm0.5$\,s from the infuser entering the water to its removal. Staged heating requires two such intervals: heater activation to food contact, then contact to heater deactivation. Success requires completing each sequence in order within its timing constraints. Each task uses 50 demonstrations and 20 evaluation trials per method. Event timestamps are annotated in evaluation videos with the same timing criteria for all methods.

\begin{table}[!htb]
\centering
\caption{\textbf{Real-robot manipulation on Galaxea A1.} Success (\%, $\uparrow$); Avg. weights all five tasks equally. Reactive denotes the standard backbone policy without persistent memory. Gains are TFP minus Reactive, in percentage points (pp).}
\label{tab:real_world_results}
\label{tab:real_world_memory}
\small
\setlength{\tabcolsep}{3pt}
\renewcommand{\arraystretch}{1.20}
\begin{tabularx}{\textwidth}{@{}l*{6}{>{\raggedleft\arraybackslash}X}@{}}
\toprule
Method & \shortstack[r]{Object\\swapping} & \shortstack[r]{Counting\\pick-place} & \shortstack[r]{Water\\pouring} & \shortstack[r]{Tea\\steeping} & \shortstack[r]{Staged\\heating} & \textbf{Avg.} \\
\midrule
\multicolumn{7}{@{}l}{\emph{Diffusion Policy}} \\
Reactive & 10.0 & 30.0 & 20.0 & 15.0 & 10.0 & 17.0 \\
\rowcolor{TableTint}
\textbf{TFP} \textit{(ours)} & \best{60.0} & \best{75.0} & \best{60.0} & \best{65.0} & \best{60.0} & \best{64.0} \\
\gainlabel{Gain (pp)} & \gain{50.0} & \gain{45.0} & \gain{40.0} & \gain{50.0} & \gain{50.0} & \gain{47.0} \\
\midrule
\multicolumn{7}{@{}l}{\emph{$\pi_{0.5}$}} \\
Reactive & 15.0 & 40.0 & 30.0 & 25.0 & 20.0 & 26.0 \\
\rowcolor{TableTint}
\textbf{TFP} \textit{(ours)} & \best{75.0} & \best{90.0} & \best{70.0} & \best{75.0} & \best{70.0} & \best{76.0} \\
\gainlabel{Gain (pp)} & \gain{60.0} & \gain{50.0} & \gain{40.0} & \gain{50.0} & \gain{50.0} & \gain{50.0} \\
\bottomrule
\end{tabularx}
\tabnote{20 trials per task and method. Bold compares Reactive and TFP within each backbone.}
\end{table}

TFP improves all five tasks, with mean gains of 47 and 50 percentage points for Diffusion Policy and $\pi_{0.5}$, respectively (Table~\ref{tab:real_world_results}). Both backbones gain 50 points on tea steeping and staged heating, extending the timing results to physical duration constraints. See Appendix~\ref{app:real_world_details} for details.

\subsection{Memory Beyond Explicit Timing Tasks}
\label{sec:main_benchmark_results}

We evaluate $\pi_{0.5}$ with TFP on LIBERO~\citep{liu2023libero} and visually perturbed LIBERO-plus~\citep{fei25libero-plus} (Table~\ref{tab:libero_main_results}). Reproduced memory variants share demonstrations, execution settings, and evaluation protocols. Standard LIBERO uses three training seeds and 50 rollouts per task and seed.

\begin{table}[!ht]
\centering
\caption{\textbf{General manipulation and visual robustness.} Success (\%, $\uparrow$); reproduced variants report means over three training seeds. Bold marks column maxima across listed methods, including ties. Gains compare TFP with the corresponding reactive $\pi_{0.5}$ baseline (pp).}
\label{tab:libero_main_results}
\label{tab:libero_plus_main_results}
\label{tab:memory_dynamics_ablation}
{\small
\setlength{\tabcolsep}{2pt}
\renewcommand{\arraystretch}{1.24}
\begin{tabularx}{\textwidth}{@{}>{\raggedright\arraybackslash}X*{9}{r}@{}}
\toprule
\multirow{2}{*}{\textbf{Policy}} & \multicolumn{5}{c}{\textbf{LIBERO}} & \multicolumn{4}{c}{\textbf{LIBERO-plus}} \\
\cmidrule(lr){2-6}\cmidrule(lr){7-10}
 & \textbf{Spatial} & \textbf{Object} & \textbf{Goal} & \textbf{Long-10} & \textbf{Avg.} & \textbf{Noise} & \textbf{Light} & \textbf{Backgr.} & \textbf{Avg.} \\
\midrule
\multicolumn{10}{@{}l}{\emph{Published results}} \\
Reactive $\pi_{0.5}$ (OpenPI) & 98.8 & 98.2 & 98.0 & 92.4 & 96.9 & -- & -- & -- & -- \\
HAMLET (3B) & 99.0 & \best{100.0} & 99.2 & 92.2 & 97.6 & -- & -- & -- & -- \\
AVA-VLA~\citep{xiao2025avavla} (7.3B) & 97.4 & 99.4 & 97.4 & \best{97.6} & 98.0 & 78.0 & 95.5 & 88.9 & 87.5 \\
MemoryVLA (7.3B) & 98.4 & 98.4 & 96.4 & 93.4 & 96.7 & -- & -- & -- & -- \\
\midrule
\multicolumn{10}{l}{\emph{Reproduced $\pi_{0.5}$ variants (3.3B)}} \\
Reactive (LIBERO-plus) & -- & -- & -- & -- & -- & 85.2 & 93.9 & 95.1 & 91.4 \\
GRU & 99.0 & 98.4 & 95.8 & 89.1 & 95.6 & 85.9 & 91.5 & 96.1 & 91.1 \\
S4D & 98.5 & 98.4 & 97.0 & 92.5 & 96.6 & 82.6 & 95.0 & 92.3 & 90.0 \\
\rowcolor{TableTint}
\textbf{TFP} \textit{(ours)} & \best{99.5} & 99.0 & \best{99.3} & 96.9 & \best{98.7} & \best{88.5} & \best{96.1} & \best{96.6} & \best{93.7} \\
\gainlabel{Gain vs. Reactive (pp)} & \gain{0.7} & \gain{0.8} & \gain{1.3} & \gain{4.5} & \gain{1.9} & \gain{3.3} & \gain{2.1} & \gain{1.5} & \gain{2.3} \\
\bottomrule
\end{tabularx}
\tabnote{Standard LIBERO Reactive: official \href{https://github.com/Physical-Intelligence/openpi/blob/main/examples/libero/README.md}{OpenPI} results. Our variants use 50 rollouts per task and training seed. Gains use unrounded averages. Seed SDs and TFP rollout CIs: Appendix~\ref{app:libero_uncertainty}. Published policies use their reported settings; -- denotes an unreported result.}
}
\end{table}

TFP achieves 98.7\% mean LIBERO success, compared with the reported reactive baseline of 96.9\%. The largest gap is on Long-10: 96.9\% versus 92.4\% (+4.5 points), compared with gaps of 0.7--1.3 points on the other suites. On LIBERO-plus, success rises from 91.4\% to 93.7\%, leading reproduced variants in every condition. Appendices~\ref{sec:ablation_memory_dynamics} and~\ref{app:statistical-reliability} give variant settings and rollout uncertainty.

\section{Conclusion}
\label{sec:conclusion}

We identify physical time as a missing dimension of robot policy memory
driven by observation sequences. TFP retains past events in a memory state
that evolves over measured time intervals. Observations and history
determine the content and timescales of the transition, while measured
intervals determine how far the memory state advances. This state
conditions existing diffusion and flow-matching action generators.

Comparisons with recurrent and state-space memories supplied with the
same measured intervals support using elapsed time to govern memory
evolution. Memory readouts recover elapsed event time beyond episode
progress and motion information after occlusion. Clock-scaling
interventions systematically advance or delay button release and passage
entry, while memory resets and conditioning ablations support the role
of accumulated history in control. Real-robot tea steeping and staged
heating extend the evidence to physical duration constraints;
memory-dependent tasks and LIBERO show broader benefits in manipulation. When actions depend on how long ago an event occurred, memory dynamics
should follow physical time.

\label{sec:main_text_end}

\clearpage
\bibliographystyle{iclr2027_conference}
\bibliography{reference}

\clearpage
\appendix

\providecommand{\vecsym}[1]{\mathbf{#1}}

\section{Additional Method Details}
\label{app:method_details}

\subsection{Sequence Training}
\label{app:eatb_details}

The controlled timing experiments reconstruct memory from causal
prefixes. For $\pi_{0.5}$, the recurrence starts from zero and gradients
flow through the full prefix to each of four supervised queries.
Padded observations leave memory unchanged and padded actions
are excluded from the loss. LIBERO instead uses consecutive episode
segments with memory carried forward and gradients truncated between
segments, as detailed below.

For the LIBERO $\pi_{0.5}$ instantiation, each episode $e$ has a hidden state
$\vecsym{h}_n^{(e)}=\mathcal{U}_\theta(\vecsym{h}_{n-1}^{(e)},\vecsym{x}_n^{(e)},\Delta_n^{(e)})$,
initialized to $\vecsym{h}_{-1}^{(e)}=\mathbf{0}$ and reset only at episode boundaries.
At optimization step $r$, we sample $B$ active episodes and unroll each for $K$
consecutive queries. For episode $e_i$, let $n_i$ be the segment's first
query index and $k=0,\ldots,K-1$ index queries within the segment, so that
$\vecsym{x}_{i,k}=\vecsym{x}_{n_i+k}^{(e_i)}$ and
$\Delta_{i,k}=\Delta_{n_i+k}^{(e_i)}$.
Writing $\tilde{\vecsym{h}}$ for the stored state, we retrieve
$\vecsym{h}_{i,-1}=\tilde{\vecsym{h}}_{n_i-1}^{(e_i)}$ and update
\begin{equation}
\vecsym{h}_{i,k}
=
\mathcal{U}_\theta(\vecsym{h}_{i,k-1},\vecsym{x}_{i,k},\Delta_{i,k}),
\qquad k=0,\dots,K-1 .
\end{equation}
The optimization objective is
\begin{equation}
\mathcal{L}_r
=
\frac{1}{BK}
\sum_{i=1}^{B}
\sum_{k=0}^{K-1}
\ell_{\mathrm{flow}}^{(i,k)} .
\end{equation}
After the parameter update, the final hidden state is written back with gradient
truncation,
\begin{equation}
\tilde{\vecsym{h}}_{n_i+K-1}^{(e_i)}
\leftarrow
\operatorname{stopgrad}(\vecsym{h}_{i,K-1}) .
\end{equation}
This carries memory across training segments while bounding the
backpropagation length.

\subsection{Memory Initialization and Parameterization}
\label{app:memory_parameterization}

Using zero-based query indices, memory is initialized as
$\vecsym{h}_{-1}=\mathbf{0}$ at each episode boundary. The controlled
timing tasks use a bootstrap interval $\Delta_0=0.1$\,s; subsequent updates
use timestamp differences. In experimental labels, $\Delta t$ denotes
this elapsed-time input.

\paragraph{Diffusion Policy.}
The memory has 128 channels. Separate heads predict a $\tanh$-bounded
candidate and positive time constants using $\operatorname{softplus}$
with a $10^{-6}$ floor.
For the controlled Button and Passage recipes, time-constant head
weights start at zero with the default linear-layer bias; Interception
retains the default linear-layer initialization.

\paragraph{$\pi_{0.5}$.}
The memory has 256 channels and a $\tanh$-bounded candidate. For the
controlled timing tasks, a sigmoid maps time constants to
$[\tau_{\min},\tau_{\max}]$, set by the
median query interval and the longest episode duration in the training
data, respectively. The time-constant head has zero initial weights;
its biases initialize the channel time constants geometrically from
$1.01\tau_{\min}$ to $\tau_{\max}/1.01$. The memory-to-action projection
starts with its default weights scaled by 0.02 and zero bias.
These TFP implementations use the exponential update in
Equation~\ref{eq:memory_update}.

\subsection{Relation to Canonical LTC Dynamics}
\label{app:ltc_integration}
\label{app:old_update_relation}

Canonical LTC dynamics~\citep{hasani2021ltc} take the form
\begin{equation}
\frac{d\vecsym{h}}{du}
=-\vecsym{\tau}_0^{-1}\odot\vecsym{h}
 +\vecsym{f}_{\theta}(\vecsym{h},\vecsym{x})\odot(\vecsym{a}-\vecsym{h}),
\label{eq:ltc_dynamics}
\end{equation}
with positive baseline time constants and a nonnegative drive. Writing
\begin{equation}
\vecsym{\lambda}=\vecsym{\tau}_0^{-1}+\vecsym{f}_{\theta},
\qquad
\vecsym{\mu}=\frac{\vecsym{f}_{\theta}\odot\vecsym{a}}{\vecsym{\lambda}},
\label{eq:ltc_relaxation_parameters}
\end{equation}
gives the relaxation form
\begin{equation}
\frac{d\vecsym{h}}{du}=\vecsym{\lambda}\odot(\vecsym{\mu}-\vecsym{h}).
\label{eq:ltc_relaxation}
\end{equation}
For frozen coefficients its exact update is
\begin{equation}
\vecsym{h}^{+}=e^{-\delta\vecsym{\lambda}}\odot\vecsym{h}^{-}
 +(1-e^{-\delta\vecsym{\lambda}})\odot\vecsym{\mu}.
\label{eq:ltc_local_solution}
\end{equation}
The interpolation has the same form as TFP when
$\vecsym{\tau}_n=\vecsym{\lambda}^{-1}$ and
$\widehat{\vecsym{h}}_n=\vecsym{\mu}$. However, canonical LTC couples these
quantities through the same drive, whereas TFP predicts them independently.
TFP shares this local relaxation form, with separate heads for candidate content
and time constants.

As in CfC~\citep{hasani2022cfc}, content and temporal modulation depend on
inputs and prior state. TFP uses this dependence to retain task-relevant events
in persistent memory, evolve that memory over measured physical intervals,
and condition diffusion and flow-matching action generators.

\section{Properties of the Memory Update}
\label{app:memory_interpretation}

The memory dynamics in Equation~\ref{eq:memory_update}
admit a local relaxation interpretation. We analyze the update with its predicted
candidate and time constants fixed over the interval of a single policy query.

\subsection{Local Relaxation and Interval Composition}
\label{app:time_consistency}
\label{app:elapsed_time_consistency}

For fixed candidate $\vecsym{\mu}$, positive rate $\vecsym{\lambda}$,
and initial state $\vecsym{h}$, the solution of
$d\vecsym{y}/du=\vecsym{\lambda}\odot(\vecsym{\mu}-\vecsym{y})$ is
\begin{equation}
\Phi_{\delta}(\vecsym{h})
=\vecsym{\mu}
 +e^{-\delta\vecsym{\lambda}}\odot(\vecsym{h}-\vecsym{\mu}).
\label{eq:fixed_relaxation_map}
\end{equation}
Setting $\vecsym{\mu}=\widehat{\vecsym{h}}_n$ and
$\vecsym{\lambda}=\vecsym{\tau}_n^{-1}$ recovers TFP's update.
At zero duration, $\Phi_0(\vecsym{h})=\vecsym{h}$. For $\delta\geq 0$,
each coordinate interpolates between the initial state and the candidate,
with distance
\begin{equation}
\left|[\Phi_{\delta}(\vecsym{h})]_j-\mu_j\right|
=e^{-\delta\lambda_j}|h_j-\mu_j|.
\label{eq:fixed_relaxation_distance}
\end{equation}
Since $\lambda_j>0$, this distance decreases monotonically with duration
unless $h_j=\mu_j$, in which case the coordinate remains unchanged.
Direct substitution also gives
\begin{equation}
\Phi_{\delta_2}(\Phi_{\delta_1}(\vecsym{h}))
=\Phi_{\delta_1+\delta_2}(\vecsym{h}).
\label{eq:fixed_relaxation_composition}
\end{equation}
Splitting an interval leaves the result unchanged if both parts use
the same candidate and rate.

For scalar retention, the exponential form follows from
this composition condition. If a continuous function
$k:[0,\infty)\rightarrow(0,1]$ satisfies $k(0)=1$ and
$k(\delta_1+\delta_2)=k(\delta_1)k(\delta_2)$, then $\log k$ is continuous
and additive, so $\log k(\delta)=-\lambda\delta$ for some $\lambda\geq 0$.
Consequently, $k(\delta)=e^{-\lambda\delta}$.

TFP recomputes the candidate and rate from the observation and memory at
each query, then holds them fixed over the local integration interval.

\paragraph{Temporal inductive bias and unseen intervals.}
Identity at zero duration, monotonic approach to a candidate, and interval
composition constrain how elapsed time affects the transition. A generic
learned gate that receives elapsed time as a feature does not structurally
enforce these properties and must learn its temporal response from data.
For a fixed candidate and positive time constants, TFP's transition remains
defined for every nonnegative interval, including intervals outside the
training range. Each coordinate stays between its initial value and its
candidate, and converges to the candidate as the interval tends to infinity.
This specifies the transition's behavior beyond observed intervals without
requiring the model to learn those constraints from examples at every
duration. Table~\ref{tab:core_structure_ablations} compares this exponential
update with a learned temporal gate.

\subsection{Duration Sensitivity}
\label{app:duration_sensitivity}

Differentiating Equation~\ref{eq:fixed_relaxation_map} with respect to the
interval gives
\begin{equation}
\frac{\partial\Phi_{\delta}(\vecsym{h})}{\partial\delta}
=\vecsym{\lambda}\odot e^{-\delta\vecsym{\lambda}}
 \odot(\vecsym{\mu}-\vecsym{h}).
\label{eq:duration_sensitivity}
\end{equation}
The state changes with duration when it differs from the candidate, with the
rate controlling how quickly that difference decays. At equilibrium,
$\vecsym{h}=\vecsym{\mu}$, duration sensitivity vanishes.

\section{Relation to Alternative Temporal Models}
\label{app:baseline_comparison}

\subsection{Recurrent and Continuous-Time Models}
\label{sec:gru_comparison}

A GRU~\citep{cho2014learning} or LSTM supplied with $\Delta_n$ can learn duration-dependent transitions
from temporal inputs. Without that input, identical observation sequences
produce the same recurrent trajectory even if they span different durations.
TFP assigns elapsed duration a specific role in the transition: it sets the
integration interval of the learned relaxation dynamics.

Time-aware recurrent models also handle irregular observations.
GRU-D uses elapsed time in learned input and hidden-state
decays~\citep{che2018grud}, while ODE-RNNs and Latent ODEs integrate hidden
dynamics over irregular observation intervals~\citep{rubanova2019latentode}.
These models establish that elapsed time can shape hidden-state evolution
beyond serving as an additional input.

A continuous-time relaxation baseline can also use a constant learned rate,
\begin{equation}
\frac{d\vecsym{h}}{du}
=\vecsym{\lambda}_0\odot
 (\vecsym{q}_{\theta}(\vecsym{h},\vecsym{x})-\vecsym{h}),
\qquad \vecsym{\lambda}_0>\mathbf{0}.
\end{equation}
Comparing this baseline with TFP tests the value of adapting the update rate
to state and input while preserving history-dependent candidate content.

\subsection{State-Space Models and Computational Cost}
\label{app:ssm_comparison}
\label{app:computational_overhead}

Continuous-time state-space models~\citep{gu2021s4} can likewise represent transitions that
depend on elapsed time, and input-dependent state-space models provide
additional adaptive mechanisms~\citep{gu2023mamba}. Our manipulation comparison includes S4D
alongside recurrent variants within the same
action-conditioning setup (Appendix~\ref{sec:ablation_memory_dynamics}).
Mamba's learned discretization step controls selection and is not itself
a measured physical interval.

Evaluating the candidate and time-constant heads costs
$\mathcal{O}(d_h(d_x+d_h))$ per query, followed by an
$\mathcal{O}(d_h)$ elementwise update. Projecting the memory costs
$\mathcal{O}(d_h d_c)$, with additional
adapter work depending on the decoder. Here $d_x$ and $d_c$ denote the
observation-feature and decoder-conditioning widths, respectively.
Each active episode requires
$\mathcal{O}(d_h)$ persistent state, independent of episode length.
These costs are incurred at the existing policy queries.

\begingroup
\setlength{\floatsep}{8pt plus 2pt minus 2pt}
\setlength{\textfloatsep}{10pt plus 2pt minus 2pt}
\setlength{\intextsep}{8pt plus 2pt minus 2pt}
\makeatletter
\setlength{\@fptop}{0pt}
\setlength{\@fpsep}{10pt}
\setlength{\@fpbot}{0pt plus 1fil}
\makeatother
\section{Additional Experimental Details}
\label{app:experimental_details}
\subsection{Controlled Timing Protocols}
\label{app:timing_protocol}

\paragraph{Model configurations.}
Table~\ref{tab:timing_architecture} summarizes the controlled-task models,
trained separately for each task. Diffusion Policy uses a compact visual
architecture~\citep{chi2023diffusion}, trained from scratch: convolutional
stages of 16, 32, and 64 channels, $3\times3$ pooling, and a
128-dimensional projection. Its denoiser has three Transformer blocks
of width 128, four attention heads, and feed-forward width 512.
The $\pi_{0.5}$ models freeze the vision-language backbone and train the
action expert, memory, and conditioning projections. They use one RGB camera and a continuous state
token. Both backbones retain current-observation conditioning and keep
memory fixed throughout each query's solver steps.

\begin{table}[!htb]
\centering
\caption{TFP architecture and sampling settings for the controlled timing tasks.}
\label{tab:timing_architecture}
\small
\setlength{\tabcolsep}{5pt}
\renewcommand{\arraystretch}{1.08}
\begin{tabularx}{\textwidth}{@{}lXX@{}}
\toprule
Setting & Diffusion Policy & $\pi_{0.5}$ \\
\midrule
RGB resolution & $96\times96$ & $224\times224$ \\
Observation summary $d_x$ & $2\times128$ (vision, state) & $3\times256$ (vision, state, text) \\
Memory width $d_h$ & 128 & 256 \\
Input to each memory head & $384=256+128$ & $1024=768+256$ \\
Candidate head & Linear + $\tanh$ & Linear + RMS norm + $\tanh$ \\
Time-constant head & Linear + softplus + $10^{-6}$ & Linear + bounded sigmoid \\
Memory projection & $128\to128$, AdaLN & $256\to1024$, AdaRMS \\
Action horizon & 16 & 10 \\
Evaluation sampler & 50-step DDIM, $\eta=0$ & 20-step Euler flow solver \\
\bottomrule
\end{tabularx}
\tabnote{For $\pi_{0.5}$, each modality projection is RMS-normalized before
concatenation. Time-constant ranges and initialization are given in
Appendix~\ref{app:memory_parameterization}. DP uses 100 training diffusion
timesteps with a cosine noise schedule and predicts clean actions
($x_0$). The short-history baseline uses two frames.}
\end{table}

\paragraph{Baseline memory implementations.}
Recurrent alternatives replace the memory core while retaining each
backbone's observation projections and action-conditioning pathway.
Their memory outputs have width 128 for Diffusion Policy and 256 for
$\pi_{0.5}$.

For \textbf{GRU+$\Delta t$}, Diffusion Policy adds a learned affine
projection of the interval in seconds to the observation features before
the GRU update. The $\pi_{0.5}$ implementation instead concatenates
$\log(1+\Delta t/(0.1\,\mathrm{s}))$ with the observation and recurrent
features used by the reset/update gates and candidate. \textbf{Plain
GRU} is trained separately without this explicit time input at training or evaluation.

\textbf{S4D+$\Delta t$} uses a diagonal complex-valued state-space
recurrence with zero-order-hold discretization. Measured intervals scale
each channel's learned nominal step by $\Delta t/(0.1\,\mathrm{s})$.
A learned readout, input skip connection, and gated output projection
produce the memory condition. For both Diffusion Policy and $\pi_{0.5}$,
\textbf{plain S4D} is trained independently with a constant 0.1-second
input interval throughout training and evaluation.

\textbf{LTC+$\Delta t$} receives elapsed seconds directly in its state
updates. Diffusion Policy uses the \texttt{ncps} 1.0.1 cell;
$\pi_{0.5}$ uses the corresponding JAX implementation. LTC uses dense
sensory and recurrent conductances with positive parameter constraints
and six semi-implicit integration steps per observation interval.
We compare LTC and TFP under the same backbone and action-conditioning
interface. TFP uses the exponential update in Equation~\ref{eq:memory_update}.

\paragraph{Temporal inputs.}
TFP starts from zero memory with a common 0.1-second first interval. For both backbones, fixed-$\Delta t$ evaluation replaces each input interval with 0.1 seconds while keeping the trained weights fixed. At 10\,Hz, fixed and measured intervals coincide.

\paragraph{Training data.}
Both Diffusion Policy and $\pi_{0.5}$ use fixed 0.1-second observation
intervals (10\,Hz) during training, without mixed-cadence sampling
augmentation; the other observation schedules are used at evaluation.
The main controlled timing comparison uses three training seeds.
For DP, Button uses 400/100 training/validation demonstrations; Passage
uses 400/100 original plus 128/32 recovery demonstrations. Interception
uses 60/12 motion profiles, formed from 10/two scenes crossed with two
directions and three speeds.

\paragraph{Observation schedules and execution.}
Reported capture intervals are 0.1 seconds (10\,Hz), 0.2 seconds (5\,Hz), uniform draws from $\{0.05,0.10,0.15,0.20\}$ seconds (jitter), or draws from $\{0.10,0.20,0.40\}$ seconds with probabilities $\{0.6,0.2,0.2\}$ (dropout). A captured observation replaces the action queue; missing captures create no extra memory updates. Timeouts are 12, 22, and 6 seconds for button, passage, and interception.

\paragraph{Physical success criteria.}
Button success requires a single continuous press lasting 3.0--3.5 seconds followed by retraction above 0.30 meters. Passage presents a 0.6-second cue at a randomized onset and permits crossing during $[t_{\mathrm{cue}}+6,t_{\mathrm{cue}}+9]$; the robot must clear the window and reach the goal without collision. Interception evaluations use target speeds of 0.08, 0.12, and 0.16\,m/s in two directions under each observation schedule. The target disappears at 0.55 seconds for the remainder of the six-second episode, and the robot holds until 0.6 seconds. Success requires legal first contact with pre-collision planar velocity mismatch at most 0.04\,m/s, sustained for at least 0.1 seconds.

\paragraph{Statistics.}
Cross-schedule means weight the four schedules equally and are computed from unrounded success counts.
The ablations in Tables~\ref{tab:core_structure_ablations}
and~\ref{tab:simulation_ablations} report three-seed means.

\subsection{Memory-Structure Ablations}
\label{app:tau_ablation}

\begin{table}[!htb]
\centering
\caption{Additional Diffusion Policy ablations. Success (\%, $\uparrow$), reported as mean $\pm$ sample SD over three training seeds. Each seed's result averages the four observation schedules equally.}
\label{tab:simulation_ablations}
\label{tab:tau_structure_results}
\label{tab:memory_retention}
\label{tab:structure_complete}
\label{tab:memory_mechanisms}
\small
\setlength{\tabcolsep}{12pt}
\renewcommand{\arraystretch}{1.12}
\begin{tabular}{@{}lrr@{}}
\toprule
Variant & Button & Passage \\
\midrule
\rowcolor{TableTint}
\textbf{Full TFP} & $\mathbf{81.2 \pm 1.2}$ & $\mathbf{97.3 \pm 1.0}$ \\
\addlinespace[3pt]
$\tau$: observation & $68.3 \pm 1.5$ & $88.5 \pm 1.0$ \\
$\tau$: memory & $49.3 \pm 1.5$ & $94.0 \pm 1.0$ \\
$\tau$: static & $58.5 \pm 1.3$ & $82.7 \pm 1.3$ \\
\addlinespace[3pt]
No memory condition & $0.5 \pm 0.5$ & $3.7 \pm 0.8$ \\
Reset each observation & $0.3 \pm 0.6$ & $6.2 \pm 0.8$ \\
\addlinespace[3pt]
Shared scalar $\tau$ & $60.7 \pm 1.3$ & $93.0 \pm 1.0$ \\
Rational retention & $50.7 \pm 1.3$ & $93.0 \pm 1.0$ \\
\bottomrule
\end{tabular}
\end{table}

In Table~\ref{tab:core_structure_ablations}, static time constants retain only a learned bias, and the learned $\Delta t$ gate uses a sigmoid head on $[\vecsym{x}_n;\vecsym{h}_{n-1};\Delta_n]$. Observation features, memory width, and action-conditioning pathways are held fixed within each comparison.

The Diffusion Policy candidate and static-time-scale comparisons share task data, training budgets, and common-parameter initialization with the time-scale-input variants in Table~\ref{tab:structure_complete}. The latter restrict the time-constant head to observation or memory inputs, or use learned, input-independent time constants. These comparisons use measured intervals.

\subsection{Accumulated Memory and Action Conditioning}
\label{app:memory_retention}

Table~\ref{tab:memory_retention} compares full TFP with two separately trained ablations. \emph{No memory conditioning} removes the recurrent contribution to the decoder condition. \emph{Reset memory each observation} retains the memory update and measured interval but starts every update from zero, eliminating information carried across observations. Both preserve current-observation and diffusion-timestep conditioning. The ablations apply throughout training and evaluation under the task-specific protocols in Appendix~\ref{app:timing_protocol}. Shape-compatible parameters share initialization.

Each task uses 50 paired test scenes per schedule and seed. Results average the four schedules equally within each of three training seeds, then average across seeds.

\subsection{Update Rule}
\label{app:memory_mechanisms}

Table~\ref{tab:memory_mechanisms} compares two update variants using the task-specific protocols. Results average four schedules and three training seeds, using 50 scenes per task, schedule, and seed. \emph{Shared scalar $\tau$} predicts one time constant for all memory channels. \emph{Rational retention} replaces $\exp(-\Delta_n/\tau_{n,j})$ by $(1+\Delta_n/\tau_{n,j})^{-1}$ in each channel $j$, retaining measured time and the candidate update.

\subsection{Frozen-Weight Interventions}
\label{app:clock_interventions}

\begin{table}[!htb]
\centering
\caption{\textbf{Time-input and memory-reset interventions.} Success (\%, $\uparrow$). Each intervention uses the same trained weights as its corresponding unmodified policy. Fixed-input results average four query schedules; clock scaling uses jitter; memory resets use 10\,Hz. }
\label{tab:core_interventions}
\label{tab:pi05_clock}
\label{tab:memory_reset_intervention}
\small
\setlength{\tabcolsep}{3pt}
\renewcommand{\arraystretch}{1.04}
\begin{tabularx}{\textwidth}{ll*{6}{>{\raggedleft\arraybackslash}X}}
\toprule
 & & \multicolumn{2}{c}{\textbf{Button Press}} & \multicolumn{2}{c}{\textbf{Passage}} & \multicolumn{2}{c}{\textbf{Interception}} \\
\cmidrule(lr){3-4}\cmidrule(lr){5-6}\cmidrule(lr){7-8}
Backbone & Memory & Normal & Fixed & Normal & Fixed & Normal & Fixed \\
\midrule
\multicolumn{8}{l}{\emph{(a) Fixed time input: $\Delta t\rightarrow 0.1$\,s}} \\
\rowcolor{TableTint}
DP & \textbf{TFP} & \best{81.2} & 25.5 & \best{97.3} & 32.8 & \best{53.9} & 31.9 \\
\rowcolor{TableTint}
$\pi_{0.5}$ & \textbf{TFP} & \best{93.5} & 27.3 & \best{86.7} & 35.7 & \best{39.7} & 32.2 \\
\midrule
\multicolumn{8}{l}{\emph{(b) Scaled time input: $\Delta t\rightarrow\alpha\Delta t$}} \\
Backbone & Memory & $1.0$ & $0.5\,/\,2.0$ & $1.0$ & $0.5\,/\,2.0$ & $1.0$ & $0.5\,/\,2.0$ \\
\rowcolor{TableTint}
DP & \textbf{TFP} & \best{76.0} & 0.0 / 0.0 & \best{98.7} & 0.0 / 0.0 & \best{56.3} & 12.7 / 33.0 \\
\rowcolor{TableTint}
$\pi_{0.5}$ & \textbf{TFP} & \best{94.7} & 0.0 / 0.0 & \best{88.0} & 0.0 / 0.0 & \best{37.1} & 18.0 / 4.0 \\
\midrule
\multicolumn{8}{l}{\emph{(c) TFP: clear memory at every query}} \\
\multicolumn{2}{l}{Backbone} & Normal & Reset & Normal & Reset & Normal & Reset \\
\rowcolor{TableTint}
\multicolumn{2}{l}{DP} & \best{88.0} & 0.0 & \best{97.3} & 0.0 & \best{55.3} & 0.0 \\
\rowcolor{TableTint}
\multicolumn{2}{l}{$\pi_{0.5}$} & \best{94.0} & 0.0 & \best{90.0} & 0.0 & \best{47.3} & 0.0 \\
\midrule
\multicolumn{8}{l}{\emph{(d) TFP: clear memory once}} \\
\rowcolor{TableTint}
\multicolumn{2}{l}{DP} & \best{88.0} & 18.0 & \best{97.3} & 40.0 & \best{55.3} & 2.3 \\
\rowcolor{TableTint}
\multicolumn{2}{l}{$\pi_{0.5}$} & \best{94.0} & 2.0 & \best{90.0} & 26.0 & \best{47.3} & 5.0 \\
\bottomrule
\end{tabularx}
\par\vspace{2pt}
{\footnotesize Single resets occur at the first query at or after 3.0\,s (button), 6.0\,s (passage), or 0.6\,s (interception), measured from episode start.}
\end{table}

Each closed-loop intervention keeps the trained weights and physical task schedule
fixed and changes either the supplied interval or stored memory. Time-input changes
act at policy queries; memory resets occur at the times stated in the
table.

Figure~\ref{fig:clock_scaling} complements the success rates in
Table~\ref{tab:core_interventions}(b) with median Button and Passage event
times under the same clock scaling. With $\pi_{0.5}$, halving the supplied intervals lengthens
button contact in all 50 paired trials, whereas doubling them shortens it
in all 50. Both backbones show later passage entry with halved intervals and
earlier entry with doubled intervals in all 50 paired trials.

\subsection{Temporal and Motion Information in Memory}
\label{app:temporal_readout}

\begin{table}[!htb]
\centering
\caption{\textbf{Event time and motion in memory.} Linear-probe MAE ($\downarrow$) from frozen Diffusion Policy.}
\label{tab:time_evidence}
\small
\setlength{\tabcolsep}{3pt}
\renewcommand{\arraystretch}{1.12}
\begin{minipage}[t]{0.47\linewidth}\vspace{0pt}\centering
\begin{tabularx}{\linewidth}{@{}>{\raggedright\arraybackslash}Xrr@{}}
\toprule
\multicolumn{3}{@{}l}{\textbf{(a) Event duration (s)}} \\
\addlinespace[2pt]
Features & Button & Passage \\
\midrule
Current observation & 0.3336 & 1.3073 \\
Episode clock & 0.6178 & 0.8334 \\
\rowcolor{TableTint}
TFP memory & \best{0.0140} & \best{0.0373} \\
\bottomrule
\end{tabularx}
\end{minipage}\hfill
\begin{minipage}[t]{0.50\linewidth}\vspace{0pt}\centering
\begin{tabularx}{\linewidth}{@{}>{\raggedright\arraybackslash}Xrr@{}}
\toprule
\multicolumn{3}{@{}l}{\textbf{(b) Motion after occlusion (m/s)}} \\
\addlinespace[2pt]
Features & Velocity & Speed \\
\midrule
Current observation & 0.1200 & 0.0267 \\
First visible frame & 0.0279 & 0.0267 \\
Last visible frame & 0.0158 & 0.0194 \\
\rowcolor{TableTint}
TFP memory & \best{0.0101} & \best{0.0087} \\
\bottomrule
\end{tabularx}
\end{minipage}
\tabnote{Button: contact duration; Passage: time since cue. Motion is read at episode time 0.6\,s, after occlusion and before robot movement; targets are signed velocity or speed magnitude.}
\end{table}

\begin{table}[!htb]
\centering
\caption{Additional linear-readout comparisons using frozen policy representations, supplementing Table~\ref{tab:time_evidence}. MAE and RMSE are reported separately; direction is classification accuracy.}
\label{tab:time_evidence_controls}
\label{tab:temporal_readout}
\label{tab:pi05_cue_readout}
\small
\setlength{\tabcolsep}{3pt}
\renewcommand{\arraystretch}{1.12}
\begin{minipage}[t]{0.55\linewidth}\vspace{0pt}\centering
\begin{tabularx}{\linewidth}{@{}>{\raggedright\arraybackslash}Xrr@{}}
\toprule
\multicolumn{3}{@{}l}{\textbf{(a) DP: duration MAE (s)} $\downarrow$} \\
\addlinespace[2pt]
Features & Button & Passage \\
\midrule
\rowcolor{TableTint}
Observation + memory & \best{0.0121} & \best{0.0355} \\
Episode clock & 0.6178 & 0.8334 \\
\bottomrule
\end{tabularx}
\end{minipage}\hfill
\begin{minipage}[t]{0.42\linewidth}\vspace{0pt}\centering
\begin{tabularx}{\linewidth}{@{}>{\raggedright\arraybackslash}Xr@{}}
\toprule
\multicolumn{2}{@{}l}{\textbf{(b) $\pi_{0.5}$: time since cue}} \\
\addlinespace[2pt]
Features & RMSE (s) $\downarrow$ \\
\midrule
\rowcolor{TableTint}
TFP + clock & \best{0.0685} \\
GRU + clock & 0.0993 \\
History + clock & 1.5366 \\
Clock only & 1.5653 \\
\bottomrule
\end{tabularx}
\end{minipage}
\par\vspace{6pt}
\begin{tabularx}{\textwidth}{@{}l*{3}{>{\raggedleft\arraybackslash}X}@{}}
\toprule
\multicolumn{4}{@{}l}{\textbf{(c) DP Interception: after occlusion, before movement (0.6\,s)}} \\
\addlinespace[2pt]
Features & Velocity MAE (m/s) $\downarrow$ & Speed MAE (m/s) $\downarrow$ & Direction (\%) $\uparrow$ \\
\midrule
Current observation & 0.1200 & 0.0267 & 50 \\
First visible frame & 0.0279 & 0.0267 & \best{100} \\
Last visible frame & 0.0158 & 0.0194 & \best{100} \\
\rowcolor{TableTint}
Memory & \best{0.0101} & \best{0.0087} & \best{100} \\
\bottomrule
\end{tabularx}
\end{table}

For the Diffusion Policy duration readouts in Tables~\ref{tab:time_evidence} and~\ref{tab:time_evidence_controls}(a), we fit linear ridge probes to frozen full-TFP representations extracted from demonstration histories at a nominal 10\,Hz cadence. Targets are continuous hold time during button contact and elapsed time since the passage cue after it disappears. These labels supervise only the probe. The features are the current encoded observation, recurrent memory, their concatenation, or elapsed time since episode start.

Each probe uses four causal queries per episode: 128 episodes (512 queries) for fitting and 32 episodes (128 queries) for validation. Standardization uses training statistics; the ridge penalty is 1.

Table~\ref{tab:time_evidence_controls}(b) also compares readouts of time since the Passage cue from $\pi_{0.5}$ memory variants, each augmented with elapsed episode time, against the episode clock alone.

\paragraph{Motion information after occlusion.}
For Interception, we extract representations from frozen Diffusion Policy with TFP at 0.6 seconds, after the target disappears and before the robot moves. Ridge probes fit 210 prefixes from 35 scenes and are tested on 90 prefixes from 15 other scenes; all six direction--speed variants of a scene remain in the same split. Standardization and penalty selection use only fitting scenes, with five-fold grouped cross-validation. Table~\ref{tab:time_evidence_controls}(c) reports signed-velocity MAE, independently fitted speed-magnitude MAE, and direction accuracy. Single-frame controls use encoded images and robot state.

\paragraph{Navigation memory visualization.}
The demonstration in Figure~\ref{fig:navigation_memory_dynamics} has a cue at 1.7 seconds. Its layout has the readout error for time since the cue closest to the median over the 32 analyzed layouts. Channels 112 and 142 are drawn from the three largest readout contributors on the probe-fitting layouts. Heatmap rows are ordered once by their median time constant over the demonstration.

\subsection{Event-Aligned Memory in the Other Simulation Tasks}
\label{app:simulation_memory_visualizations}

Figures~\ref{fig:button_memory_dynamics} and~\ref{fig:interception_memory_dynamics} complement the Passage visualization in Figure~\ref{fig:navigation_memory_dynamics}. Both use frozen Diffusion Policy with TFP at 10\,Hz. Heatmaps show write gains $\vecsym{g}_n=\mathbf{1}-\vecsym{k}_n$ for all 128 channels, ordered once by their median time constant over the displayed sequence, with a fixed $[0,1]$ color scale. Lower panels show selected raw-memory channels $\vecsym{h}_n$; the Passage panel shows time constants.

\begin{figure}[!htb]
    \centering
    \includegraphics[width=\textwidth]{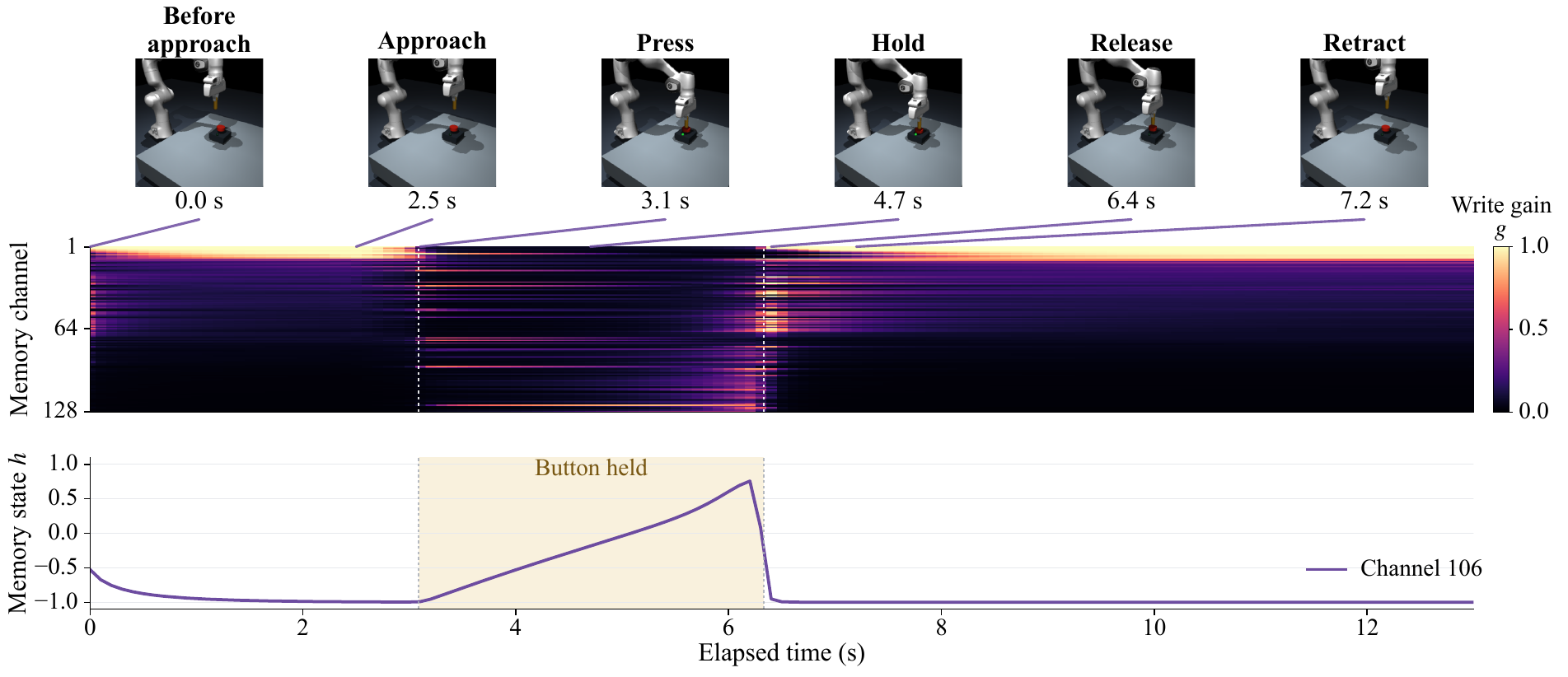}
    \caption{\textbf{Memory dynamics during Timed Button Press.} Simulation frames, write gains, and memory channel 106 are aligned in physical time for a successful sequence. Shading marks contact from 3.09 to 6.33\,s. The displayed channel rises during the hold and drops around release.}
    \label{fig:button_memory_dynamics}
\end{figure}

\begin{figure}[!htb]
    \centering
    \includegraphics[width=\textwidth]{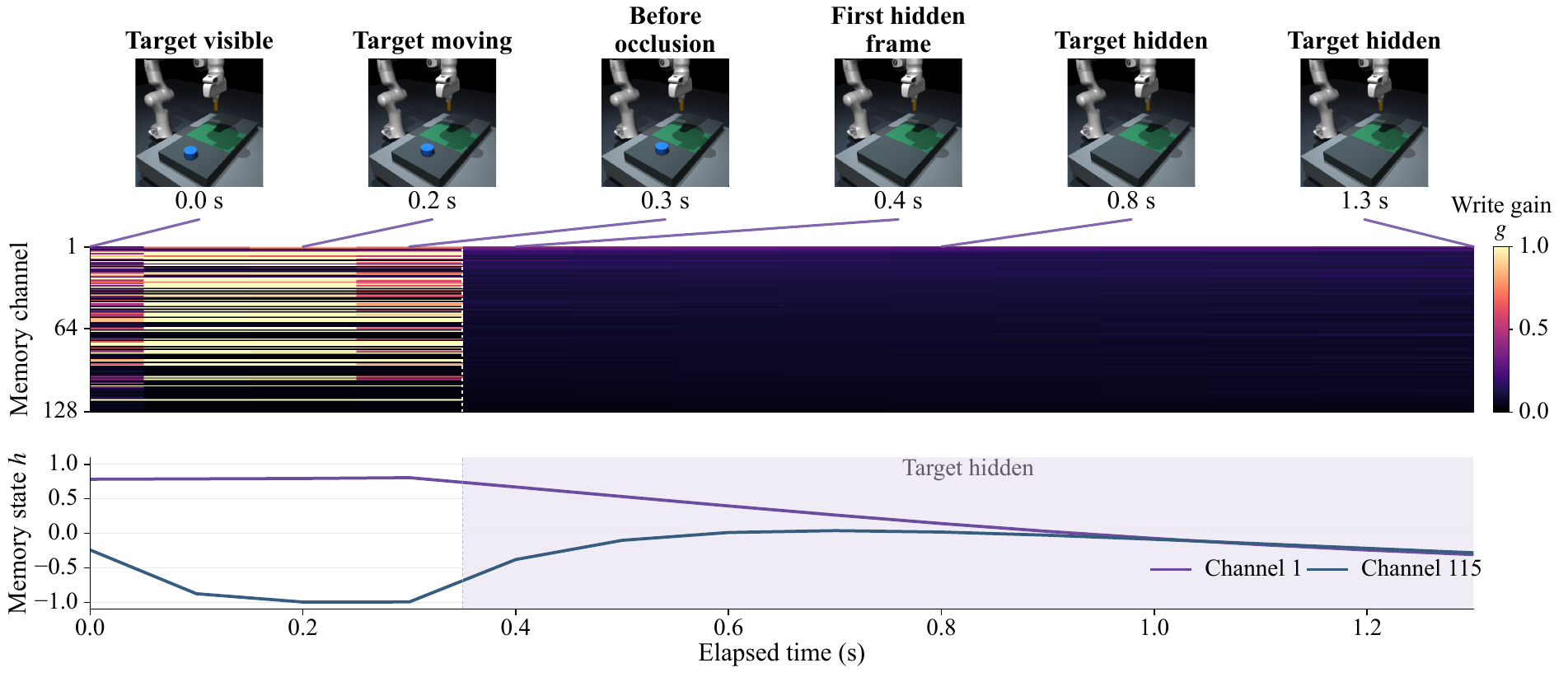}
    \caption{\textbf{Memory evolution under occlusion with a stationary robot.} The target moves at 0.16\,m/s and becomes invisible to the policy at 0.35\,s; shading marks the hidden interval. The robot remains stationary throughout this 1.3\,s observation sequence. Channels 1 and 115 evolve after occlusion while the visible observation remains unchanged. The Interception task-evaluation protocol is specified in Appendix~\ref{app:timing_protocol}.}
    \label{fig:interception_memory_dynamics}
\end{figure}

\FloatBarrier
\subsection{Manipulation Memory Ablations}
\label{sec:ablation_memory_dynamics}

The reproduced $\pi_{0.5}$ manipulation variants in
Table~\ref{tab:memory_dynamics_ablation} share demonstrations, action
horizon, optimizer, execution settings, and evaluation protocol.

\paragraph{Training cost.}
The largest $\pi_{0.5}$ runs used over 100\,GB GPU memory and approximately 80 hours on four H200 GPUs. Memory storage scales with the number of active episodes; sequence processing costs depend on the training unroll length.

\subsection{Memory Dynamics in LIBERO}
\label{app:memory_diagnostics}

Figure~\ref{fig:ltc_gain_dynamics} visualizes memory write gains during a LIBERO manipulation sequence.

\begin{figure}[!htb]
    \centering
    \includegraphics[width=\textwidth]{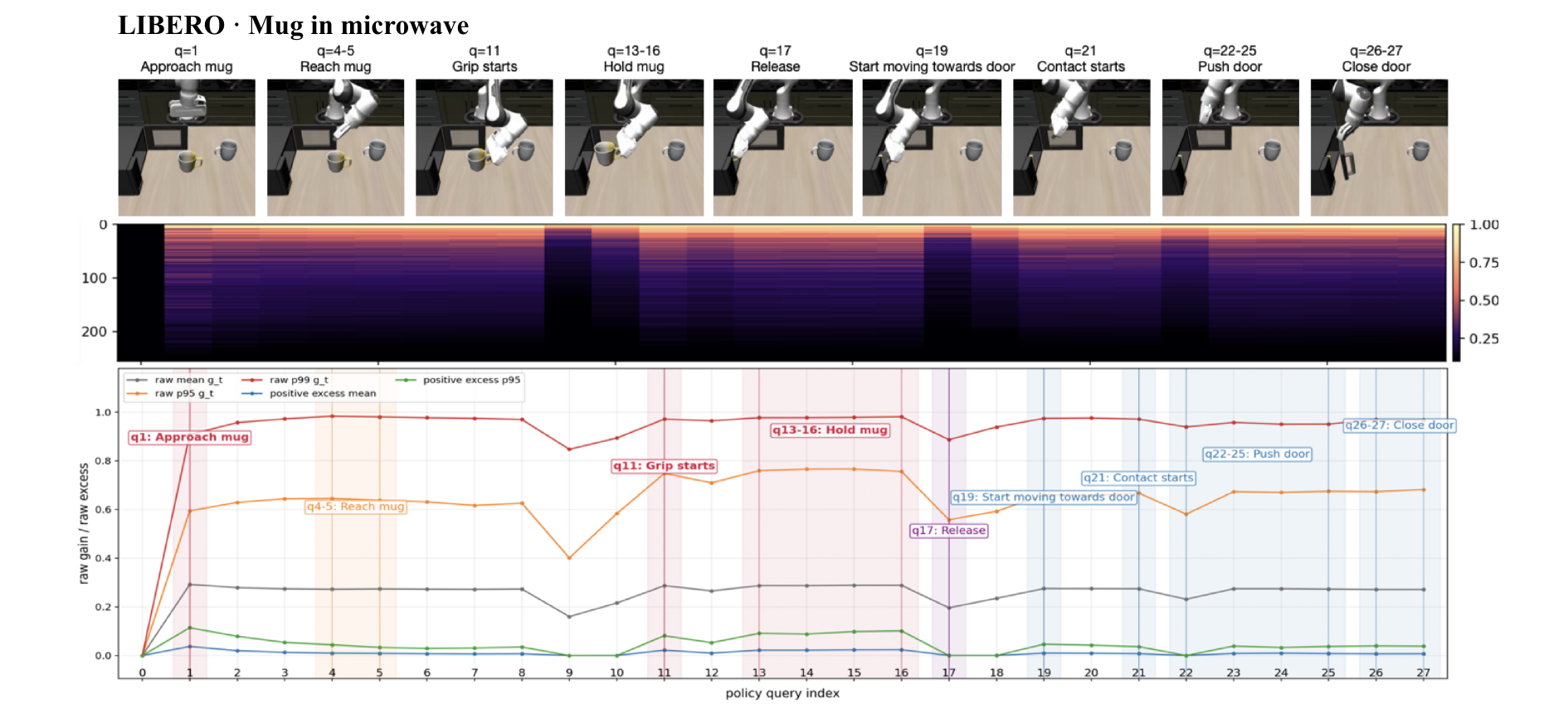}
    \caption{\textbf{Memory updates during LIBERO manipulation.} Observations (top), channel-wise write gains (middle), and gain summaries (bottom) show changes around grasping, release, and door interaction in a successful rollout. The horizontal axis indexes policy queries (denoted $t$ in the plot labels). Gains depend on both measured intervals and predicted time constants.}
    \label{fig:ltc_gain_dynamics}
\end{figure}

\FloatBarrier
\subsection{Real-World Setup and Evaluation Protocol}
\label{app:real_world_details}

We use a single-arm Galaxea A1 with a parallel gripper, a wrist-mounted
fisheye camera, and a static Intel RealSense D435 camera.

The real-robot Diffusion Policy experiments use our custom compact
visual diffusion-policy implementation described in
Appendix~\ref{app:timing_protocol}, with the same backbone architecture,
TFP memory update, and action-conditioning interface as in the controlled
simulations.

The baseline for each backbone is its standard reactive policy without
persistent memory. Each task uses 50 demonstrations for training.
Both backbones train at fixed 0.1-second query intervals (10\,Hz),
matching the controlled timing experiments. We evaluate each method
in 20 trials per task.

Tea-steeping success requires an immersion duration of $3.0\pm0.5$\,s,
measured from the infuser entering the water to its removal from the
water, followed by successful task completion. For staged heating, the
preheating interval runs from heater activation to the food contacting
the heating surface; the cooking interval runs from that contact to
heater deactivation. Both intervals must be $3.0\pm0.5$\,s, and the
robot must complete the operations in the prescribed order. Event
timestamps are annotated in evaluation videos, using the same timing
criteria for all methods.

\FloatBarrier
\section{Statistical Reliability of Rollout Results}
\label{app:statistical-reliability}

\subsection{Controlled Timing: Variation Across Training Seeds}
\label{app:timing_uncertainty}

Tables~\ref{tab:timing_main_results} and~\ref{tab:core_structure_ablations}
report mean success across three independent training seeds. Additional
Diffusion Policy ablations in Table~\ref{tab:simulation_ablations} use the
same seed-level aggregation and report mean $\pm$ sample SD. Table~\ref{tab:timing_seed_summary}
summarizes task-average success as mean $\pm$ sample standard deviation.
For each seed, we first average success equally
across 10\,Hz, 5\,Hz, jitter, and dropout, then compute the mean and
standard deviation of the three seed-level averages. Overall success in the main
text additionally weights the three tasks equally, irrespective of
their different rollout counts.

LTC+$\Delta t$ has overall success of $19.7\pm2.3$\% for Diffusion Policy
and $23.6\pm2.4$\% for $\pi_{0.5}$, averaging tasks equally within each seed.
The continuation of Table~\ref{tab:timing_seed_summary} gives its results for each
observation schedule as mean $\pm$ sample SD and rollout 95\% confidence intervals.

\begin{table}[!htbp]
\centering
\caption{\textbf{Variation across training seeds.} Task-average success (\%), reported as mean $\pm$ sample SD over three independent training seeds.}
\label{tab:timing_seed_summary}
\small
\setlength{\tabcolsep}{3pt}
\renewcommand{\arraystretch}{1.16}
\begin{tabularx}{\textwidth}{@{}l*{6}{>{\raggedleft\arraybackslash}X}@{}}
\toprule
 & \multicolumn{3}{c}{Diffusion Policy} & \multicolumn{3}{c}{$\pi_{0.5}$} \\
\cmidrule(lr){2-4}\cmidrule(lr){5-7}
Memory & Button & Passage & Intercept. & Button & Passage & Intercept. \\
\midrule
Reactive & $0.0 \pm 0.0$ & $2.3 \pm 0.3$ & $10.6 \pm 0.1$ & $0.0 \pm 0.0$ & $9.3 \pm 0.6$ & $18.3 \pm 0.3$ \\
History & $0.0 \pm 0.0$ & $2.2 \pm 0.3$ & $29.1 \pm 0.5$ & $5.5 \pm 0.0$ & $2.8 \pm 0.6$ & $35.1 \pm 0.2$ \\
GRU & $19.7 \pm 1.3$ & $37.8 \pm 0.3$ & $25.3 \pm 1.5$ & $10.5 \pm 1.5$ & $36.8 \pm 1.5$ & $27.1 \pm 1.5$ \\
GRU+$\Delta t$ & $21.0 \pm 0.0$ & $37.8 \pm 0.3$ & $26.1 \pm 0.6$ & $11.7 \pm 0.3$ & $37.3 \pm 0.3$ & $27.8 \pm 1.1$ \\
S4D & $25.7 \pm 2.1$ & $40.5 \pm 1.8$ & $24.8 \pm 1.5$ & $32.0 \pm 2.0$ & $24.7 \pm 2.3$ & $18.8 \pm 1.5$ \\
LTC+$\Delta t$ & $16.0 \pm 3.0$ & $20.3 \pm 2.5$ & $22.7 \pm 1.3$ & $20.2 \pm 2.8$ & $24.2 \pm 3.1$ & $26.6 \pm 1.5$ \\
S4D+$\Delta t$ & $48.2 \pm 2.1$ & $95.2 \pm 0.8$ & $42.8 \pm 1.5$ & $93.0 \pm 0.9$ & $74.5 \pm 0.9$ & $32.8 \pm 0.8$ \\
TFP fixed $\Delta t$ & $25.5 \pm 1.8$ & $32.8 \pm 1.0$ & $31.9 \pm 1.8$ & $27.3 \pm 1.5$ & $35.7 \pm 0.3$ & $32.2 \pm 1.5$ \\
\addlinespace[2pt]
\rowcolor{TableTint}
\textbf{TFP} \textit{(ours)} & $\mathbf{81.2 \pm 1.2}$ & $\mathbf{97.3 \pm 1.0}$ & $\mathbf{53.9 \pm 1.2}$ & $\mathbf{93.5 \pm 0.9}$ & $\mathbf{86.7 \pm 0.8}$ & $\mathbf{39.7 \pm 0.7}$ \\
\midrule
\multicolumn{7}{l}{\emph{Memory-structure ablations}} \\
Candidate w/o history & $42.7 \pm 1.3$ & $8.0 \pm 1.0$ & $44.2 \pm 0.8$ & $12.0 \pm 2.0$ & $65.7 \pm 1.5$ & $31.7 \pm 0.8$ \\
Static time constants & $58.5 \pm 1.3$ & $82.7 \pm 1.3$ & $44.8 \pm 0.7$ & $64.3 \pm 1.5$ & $13.3 \pm 8.3$ & $31.4 \pm 0.5$ \\
Learned $\Delta t$ gate & $14.5 \pm 1.0$ & $15.0 \pm 1.0$ & $30.5 \pm 0.9$ & $27.0 \pm 1.0$ & $37.0 \pm 1.0$ & $30.1 \pm 0.7$ \\
\bottomrule
\end{tabularx}
\end{table}

\begin{table}[!htbp]
\ContinuedFloat
\centering
\caption[]{\textbf{Variation across training seeds (continued).} LTC+$\Delta t$ results by observation schedule, with mean $\pm$ sample SD and rollout 95\% confidence intervals.}
\small
\setlength{\tabcolsep}{3pt}
\renewcommand{\arraystretch}{1.16}
\begin{tabularx}{\textwidth}{@{}ll*{4}{>{\raggedleft\arraybackslash}X}@{}}
\toprule
\multicolumn{6}{l}{\textbf{LTC+$\Delta t$: results by observation schedule}} \\
 & & \multicolumn{2}{c}{Diffusion Policy} & \multicolumn{2}{c}{$\pi_{0.5}$} \\
\cmidrule(lr){3-4}\cmidrule(lr){5-6}
Task & Schedule & Mean $\pm$ SD & 95\% CI & Mean $\pm$ SD & 95\% CI \\
\midrule
Button & 10 Hz & $18.7 \pm 3.1$ & $[13.2, 25.7]$ & $23.3 \pm 3.1$ & $[17.3, 30.7]$ \\
 & 5 Hz & $13.3 \pm 3.1$ & $[8.8, 19.7]$ & $18.7 \pm 3.1$ & $[13.2, 25.7]$ \\
 & Jitter & $20.7 \pm 3.1$ & $[15.0, 27.8]$ & $22.7 \pm 3.1$ & $[16.7, 30.0]$ \\
 & Dropout & $11.3 \pm 3.1$ & $[7.2, 17.4]$ & $16.0 \pm 2.0$ & $[11.0, 22.7]$ \\
\midrule
Passage & 10 Hz & $24.7 \pm 3.1$ & $[18.5, 32.1]$ & $26.7 \pm 3.1$ & $[20.2, 34.3]$ \\
 & 5 Hz & $18.0 \pm 2.0$ & $[12.7, 24.9]$ & $22.7 \pm 3.1$ & $[16.7, 30.0]$ \\
 & Jitter & $22.7 \pm 3.1$ & $[16.7, 30.0]$ & $28.7 \pm 3.1$ & $[22.0, 36.4]$ \\
 & Dropout & $16.0 \pm 2.0$ & $[11.0, 22.7]$ & $18.7 \pm 3.1$ & $[13.2, 25.7]$ \\
\midrule
Interception & 10 Hz & $24.3 \pm 1.5$ & $[21.6, 27.2]$ & $29.3 \pm 1.5$ & $[26.5, 32.4]$ \\
 & 5 Hz & $21.1 \pm 1.2$ & $[18.6, 23.9]$ & $25.3 \pm 1.5$ & $[22.6, 28.3]$ \\
 & Jitter & $25.4 \pm 1.7$ & $[22.7, 28.4]$ & $28.3 \pm 1.5$ & $[25.5, 31.4]$ \\
 & Dropout & $20.0 \pm 1.0$ & $[17.5, 22.7]$ & $23.3 \pm 1.5$ & $[20.7, 26.2]$ \\
\bottomrule
\end{tabularx}
\end{table}

\subsection{LIBERO Seed Variability and Rollout Uncertainty}
\label{app:libero_uncertainty}

The reproduced LIBERO and LIBERO-plus results use three training seeds.
Table~\ref{tab:libero_seed_summary} reports mean $\pm$ sample SD.
For each seed, Avg. weights the four LIBERO suites
or three LIBERO-plus perturbation conditions equally; its SD is computed
from the three seed-level averages. Statistics are rounded only after
aggregation.

Each standard LIBERO suite has 10 tasks with 50 rollouts per task and seed
for our evaluated variants: 500 per seed and
1,500 across three seeds.
Table~\ref{tab:statistical_reliability_libero} also reports suite-level
rollout 95\% confidence intervals for TFP. These quantify
finite-rollout uncertainty in each evaluated policy's success rate.
Reproduced variants share this protocol; published policies retain
their reported settings.

\begin{table}[!htbp]
\centering
\small
\setlength{\tabcolsep}{5pt}
\renewcommand{\arraystretch}{1.12}
\caption{LIBERO and LIBERO-plus success (\%): mean $\pm$ sample SD across three training seeds. The TFP CI row reports rollout 95\% confidence intervals on standard LIBERO.}
\label{tab:libero_seed_summary}
\label{tab:statistical_reliability_libero}
\begin{tabularx}{\textwidth}{@{}l*{5}{>{\raggedleft\arraybackslash}X}@{}}
\toprule
\multicolumn{6}{l}{\textbf{LIBERO}} \\
Policy & Spatial & Object & Goal & Long-10 & Avg. \\
\midrule
GRU & $99.0 \pm 0.2$ & $98.4 \pm 0.2$ & $95.8 \pm 0.3$ & $89.1 \pm 0.4$ & $95.6 \pm 0.1$ \\
S4D & $98.5 \pm 0.1$ & $98.4 \pm 0.2$ & $97.0 \pm 0.2$ & $92.5 \pm 0.3$ & $96.6 \pm 0.0$ \\
\rowcolor{TableTint}
\textbf{TFP} \textit{(ours)} & $99.5 \pm 0.1$ & $99.0 \pm 0.2$ & $99.3 \pm 0.1$ & $96.9 \pm 0.3$ & $98.7 \pm 0.2$ \\
\addlinespace[2pt]
TFP 95\% CI & $[99.0, 99.8]$ & $[98.4, 99.4]$ & $[98.8, 99.6]$ & $[95.9, 97.7]$ & -- \\
\bottomrule
\end{tabularx}
\par\vspace{7pt}
\begin{tabularx}{\textwidth}{@{}l*{4}{>{\raggedleft\arraybackslash}X}@{}}
\toprule
\multicolumn{5}{l}{\textbf{LIBERO-plus}} \\
Policy & Noise & Light & Background & Avg. \\
\midrule
Reactive & $85.2 \pm 0.4$ & $93.9 \pm 0.3$ & $95.1 \pm 0.2$ & $91.4 \pm 0.1$ \\
GRU & $85.9 \pm 0.4$ & $91.5 \pm 0.3$ & $96.1 \pm 0.3$ & $91.1 \pm 0.1$ \\
S4D & $82.6 \pm 0.4$ & $95.0 \pm 0.2$ & $92.3 \pm 0.3$ & $90.0 \pm 0.2$ \\
\rowcolor{TableTint}
\textbf{TFP} \textit{(ours)} & $88.5 \pm 0.5$ & $96.1 \pm 0.3$ & $96.6 \pm 0.2$ & $93.7 \pm 0.3$ \\
\bottomrule
\end{tabularx}
\end{table}

\FloatBarrier
\endgroup

\clearpage
\section{Task Sequences}
\label{app:task_sequences}

\begin{figure}[!ht]
\centering
\includegraphics[page=1,width=\textwidth,height=0.86\textheight,keepaspectratio]{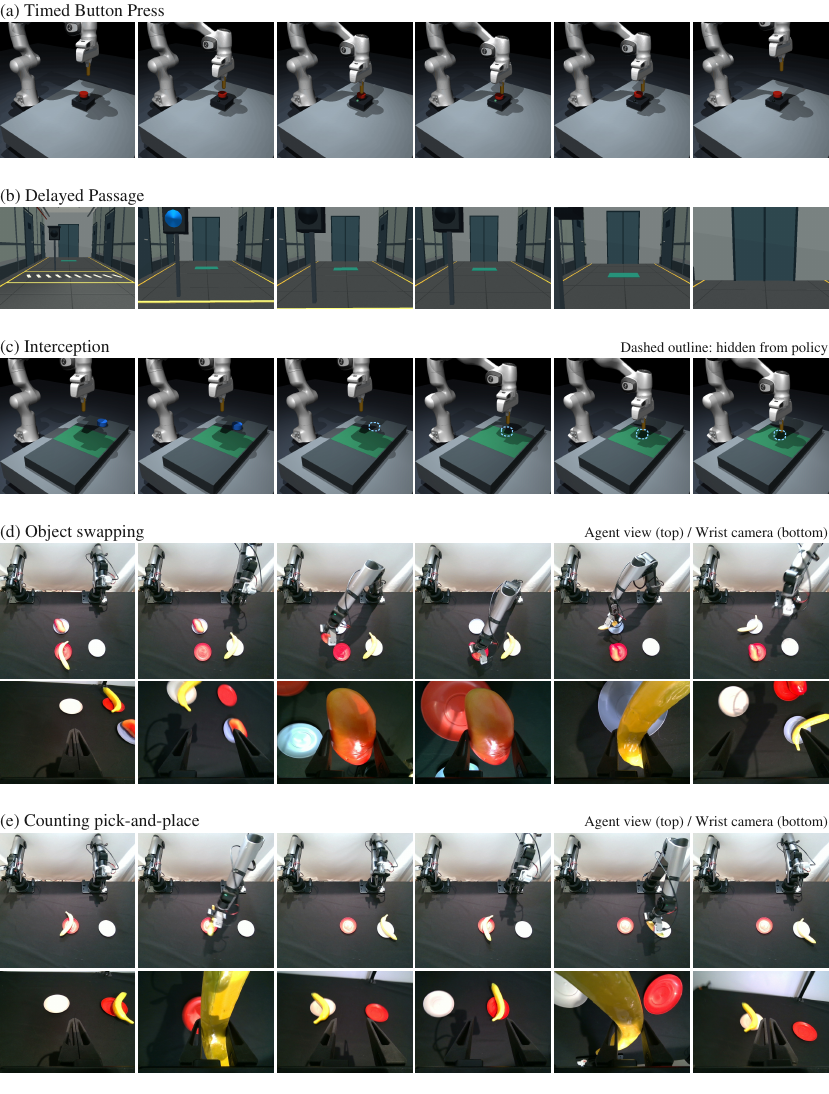}
\caption{\textbf{Qualitative task sequences} in simulation (a--c) and real-robot demonstrations (d--h, continued on the next page). Frames show action order from left to right. In the Interception expert demonstration (c), the target becomes invisible to the policy before the robot starts moving. Dashed outlines show the hidden target's true position for visualization only; they are absent from policy inputs. For real-robot tasks, each column pairs Agent view (top) and Wrist camera (bottom), preserving the model's camera input framing.}
\label{fig:task_sequences}
\label{fig:simulation_task_sequences}
\end{figure}

\clearpage
\begin{figure}[!ht]
\ContinuedFloat
\centering
\includegraphics[page=2,width=\textwidth,height=0.86\textheight,keepaspectratio]{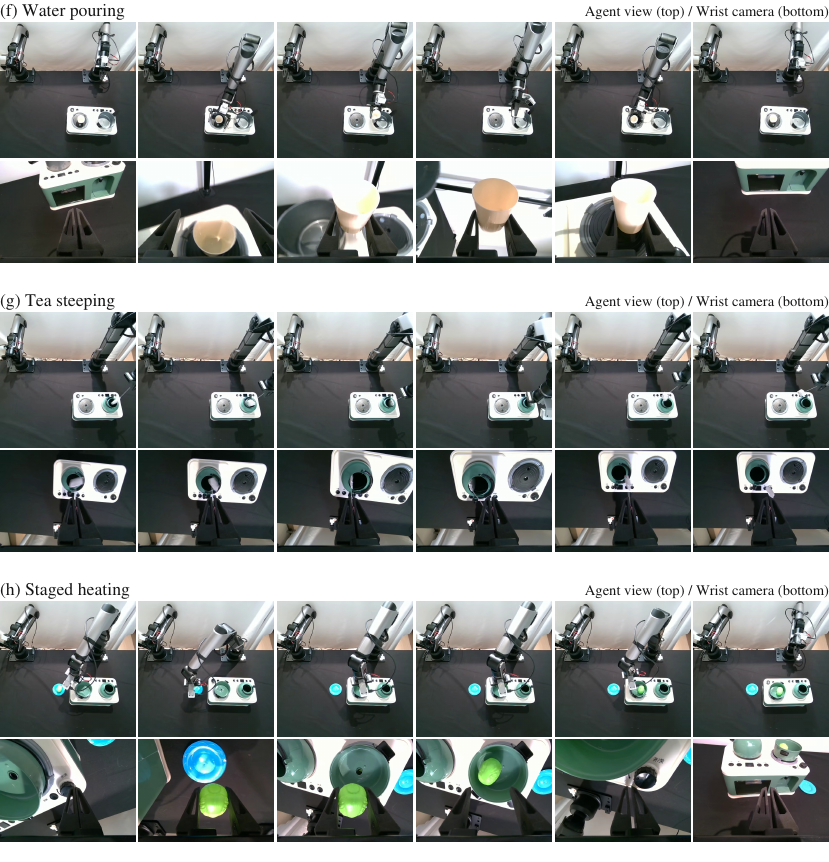}
\caption[]{\textbf{Qualitative task sequences (continued).} Paired Agent view (top) and Wrist camera (bottom) for Water pouring, Tea steeping, and Staged heating.}
\end{figure}

\end{document}